\RequirePackage{booktabs}
\documentclass[pdflatex,sn-mathphys,Numbered]{sn-jnl}

\usepackage{soul}
\usepackage{multirow}%
\usepackage{amsthm,amsmath,amssymb,amsfonts}%
\usepackage[title]{appendix}%
\usepackage{manyfoot}%
\usepackage{algpseudocode}%
\usepackage{xcolor}%
\definecolor{dr}{HTML}{FF7F0E} 
\definecolor{lr}{HTML}{FFD1A3}
\definecolor{db}{HTML}{1F77B4}
\definecolor{lb}{HTML}{85C1E9}
\theoremstyle{definition}
\newtheorem{definition}{Definition}
\newtheorem{lemma}{Lemma}
\newtheorem{theorem}{Theorem}
\newcommand{\eqdot}{\,.}

\begin{document}

\title[Article Title]{Clusterability test for categorical data}


\author[1]{\fnm{Lianyu} \sur{Hu}}\email{hly4ml@gmail.com}

\author[1]{\fnm{Junjie} \sur{Dong}}\email{jd445@qq.com}

\author[1]{\fnm{Mudi} \sur{Jiang}}\email{792145962@qq.com}

\author[2]{\fnm{Yan} \sur{Liu}}\email{yliu0414@qq.com}

\author*[1,3]{\fnm{Zengyou} \sur{He}}\email{zyhe@dlut.edu.cn}

\affil[1]{\orgdiv{School of Software}, \orgname{Dalian University of Technology}, \orgaddress{\city{Dalian}, \\ \postcode{116620}, \country{China}}}

\affil[2]{\orgdiv{School of Software Engineering}, \orgname{Dalian University}, \orgaddress{\city{Dalian}, \\ \postcode{116622}, \country{China}}}

\affil[3]{\orgname{Key Laboratory for Ubiquitous Network and Service Software of Liaoning Province}, \orgaddress{\city{Dalian}, \postcode{116620}, \country{China}}}


\abstract{The objective of clusterability evaluation is to check whether a clustering structure exists within the data set. As a crucial yet often-overlooked issue in cluster analysis, it is essential to conduct such a test before applying any clustering algorithm. If a data set is unclusterable, any subsequent clustering analysis would not yield valid results. Despite its importance, the majority of existing studies focus on numerical data, leaving the clusterability evaluation issue for categorical data as an open problem. Here we present TestCat, a testing-based approach to assess the clusterability of categorical data in terms of an analytical $p$-value. The key idea underlying TestCat is that clusterable categorical data possess many strongly associated attribute pairs and hence the sum of chi-squared statistics of all attribute pairs is employed as the test statistic for $p$-value calculation. We apply our method to a set of benchmark categorical data sets, showing that TestCat outperforms those solutions based on existing clusterability evaluation methods for numeric data. To the best of our knowledge, our work provides the first way to effectively recognize the clusterability of categorical data in a statistically sound manner.}




\maketitle

\section{Introduction}
\label{intro}
Clustering, as a multivariate analysis tool, is widely used in diverse fields such as biology and social sciences, aiming to identify a collection of homogeneous groups within the data. In the absence of a unified definition for natural clusters \cite{Kleinberg2002, Pelillo2009, Hennig2015, Thomann2015}, various properties \cite{BenDavid2008} that affect clustering quality often exhibit conflicting behavior \cite{GarzaFabre2022}, posing challenges for users when assessing the effectiveness of clustering methods. Running different clustering algorithms on identical input data can yield considerably distinct results. Typically, these approaches are evaluated within the context of specific applications \cite{VonLuxburg2012}, and a proper algorithm is selected by utilizing validity indices \cite{Arbelaitz2013} on the produced clusters. It is essential to recognize that both clustering algorithms and validity metrics hinge on the presence of underlying clustering structure. Otherwise, the functions built upon that may not output meaningful results, and utilizing these outcomes could lead to untrustworthy conclusions.

To determine whether data is clusterable before employing any clustering algorithms, different methods have been suggested to assess \textit{clusterability}~\cite{Diallo2024, Ackerman2009, Adolfsson2019, Ahmadi2022}, which aim to evaluate the extent of clustering structure present within the data. Despite the inherent ambiguity of clustering, it is evident that a data set generated from a single Gaussian distribution is unclusterable, as any partitioning on it would be spurious. Given this observation, current solutions~\cite{Adolfsson2019, Laborde2023} predominantly employ a hypothesis testing approach, specifically the multimodality tests in one dimension (Dip test~\cite{Cheng1998} and Silverman test~\cite{Silverman1981}). As the resulting $p$-value provides a probabilistic interpretation, one can determine whether the underlying structure is significant by using a significance threshold. For multivariate data sets, dimensionality reduction techniques, like PCA (Principal Component Analysis), are necessary to project the data into one-dimensional space, enabling the application of multimodality tests. In addition, several intuitive visual tools, such as VAT (Visual Assessment of Tendency)~\cite{Havens2011} and Dissimilarity plots~\cite{Hahsler2011}, can help reveal the clustering structure in data sets.

Categorical variables are pervasive in real-world data sets~\cite{Agresti2012} and serve to represent qualitative information from a variety of fields, like marital status in sociological questionnaires \cite{Couper2017} and cell types in biology \cite{Vasaikar2023}. However, all the aforementioned clusterability measures are designed specifically for numerical variables, as the dimensionality reduction techniques and the hypothesis testing procedures are not suitable for handling categorical variables. For example, in a single categorical attribute, the discrete nature of attribute values in distinct groupings, making it inappropriate to assume that the random variable follows a unimodal distribution. Furthermore, common distance metrics that several clusterability evaluation methods rely on, can not be directly applied to this type of data due to the lack of geometric properties. Consequently, in order to employ these conventional methods, one must convert categorical data into one-dimensional numerical values by utilizing dissimilarity measures explicitly designed for categorical data~\cite{Boriah2008}, which are practically ineffective (see results in~\ref{section:comparison}). Hence, identifying the existence of a categorical clustering structure remains an overlooked open challenge. When designing a clusterability evaluation method for categorical data, it is necessary to employ new significance testing strategies that differ from what have been deployed for numerical data.

Here we develop a testing-based approach, TestCat, to assess the clusterability of categorical data. Intuitively, in a clusterable categorical data set, samples in each cluster should be quite similar to each other. Accordingly, many attribute values within the cluster will be positively associated with each other. Furthermore, over-expressed attribute values in one cluster will be negatively associated with those over-expressed ones in another cluster. Hence, it can be expected that there will be many associated attribute pairs in a clusterable categorical data set. Based on the above observations, we employ the chi-squared test to measure the association for each pair of attributes and utilize the sum of chi-squared statistics of all attribute pairs as the test statistic for clusterability validation\footnote{Notably, TestCat focuses on pairwise attribute associations and may not capture complex cluster structures determined by higher-order interactions among multiple attributes where pairwise associations are absent.}. By imposing an independence assumption on attribute pairs, we can derive a $p$-value to determine the clusterability of any categorical data set. Validation studies conducted on real data sets have demonstrated the effectiveness and efficiency of our method. More importantly, the empirical results have shown that commonly used methods designed for numerical data may fail to distinguish between clusterable and unclusterable categorical data sets.

In summary, the main contributions of this work are as follows:
\begin{itemize}
	\item The problem of clusterability assessment for categorical data is conceptualized as an issue of testing association among attributes. To the best of our knowledge, this is the first attempt to develop a method specifically designed for assessing the clusterability of categorical data.
	\item A $p$-value based on chi-squared tests is derived as a validation index to determine the clusterability of categorical data. Experimental results on real categorical data sets demonstrate that the proposed method is both effective and robust, while also being comparable in efficiency to other methods.
	\item To date, how to empirically compare the performance of different clusterability evaluation methods for categorical data is still an open problem. A feasible and reasonable pipeline is presented and adopted in our empirical studies, which may serve as a practical validation strategy for future research towards this direction. 
\end{itemize}

The rest of this paper is organized as follows: Section~\ref{relatedork} provides a review of existing methods that are closely related to our topic. Section~\ref{methods} offers a detailed description of our method. Section~\ref{results} presents the results, including experiments on real data sets and further analysis. Lastly, Section~\ref{conclusions} concludes the paper.

\section{Related Work}
\label{relatedork}
Currently, there are no clusterability evaluation methods explicitly designed for categorical data. Consequently, we will provide separate overviews of clustering methods for categorical data and clusterability evaluation methods for numerical data. It is important to note that clustering algorithms for categorical data do not readily lend themselves to clusterability evaluation methods, and at the same time, the testing-based approaches used in existing clusterability evaluation methods are not well-suited for handling categorical data.

\subsection{Categorical data clustering}
Existing categorical data clustering algorithms~\cite{Naouali2020}, which include various approaches such as partition-based, model-based, density-based, and linkage-based hierarchical methods, can generally be classified into two main types based on the objective functions they employ: $k$-means-type~\cite{Liu2023} and others~\cite{Bouguessa2015,Cesario2007}. The $k$-means-type algorithms quantify each cluster by calculating the sum of the squared distances~\cite{Zhang2022} between each object and its cluster center (i.e., mode)~\cite{Xiao2019}. Additionally, some of these algorithms utilize Categorical-to-Numerical techniques~\cite{Jian2019, Bai2022}, transforming categorical data into numerical data, and then applying the original $k$-means algorithm. In contrast, other algorithms use measures like Entropy~\cite{Barbara2002, Li2004} or Category Utility~\cite{Fisher1987, Mirkin2001} to quantify a cluster. They assess the difference between the observed category distribution in a cluster and the expected distribution under a random assignment of objects to clusters.

The value of objective function is data-dependent and can vary significantly across different data sets. The numerical value itself does not offer insights into the quality of individual clusters or the overall partition. Instead, it represents a local optimum among a myriad of possible partitions. Furthermore, even when we have the global optimum values, they do not signify the presence of a clustering structure or indicate whether the clusters are obtained from a randomized data set. Notably, methods based on Category Utility merely measure the difference between the observed category distribution and the expected category distribution, failing to provide a significance-based score in terms of $p$-value. Anyway, all these objective functions can be used to evaluate the goodness of clustering results, however, they cannot be directly used for the purpose of clusterability evaluation. 

\subsection{Clusterability evaluation methods}
Without relying on any specific partitioning, standard-compliant clusterability evaluation methods have been proposed, as seen in~\cite{Adolfsson2019,Laborde2023}. Meanwhile, algorithm-dependent methods, such as those presented in~\cite{Epter1999, Liu2008}, have been excluded from consideration as they do not adequately meet the practical requirements of clusterability assessment. More specifically, the output values of the clusterability evaluation function can be used to declare data as  either clusterable or unclusterable. This goal is well-achieved by testing-based approaches that employ a $p$-value as the validation index, providing a statistically meaningful alternative to other methods.

For numerical data, testing-based clusterability evaluation methods conduct the analysis on either the original data or transformed data.  A single cluster typically arises from a homogeneous Gaussian distribution, so the presence of multiple clusters suggests a deviation from this pattern.   Accordingly, detecting multimodality in a data set with statistical methods such as Dip~\cite{Cheng1998} and Silverman~\cite{Silverman1981} serves as a proxy for clusterability assessment. Similarly, the concept that a single cluster corresponds to spatial randomness has led to the development of the Hopkins statistic~\cite{Dubes1987} and the PHI statistic~\cite{Diallo2024}.

When addressing categorical data, the approach to clusterability evaluation needs to be tailored to accommodate the unique characteristics of this data type. The test statistics mentioned earlier are not directly applicable to the original data, as categories cannot be treated as numerical values. Likewise, dimensionality reduction techniques typically used for numerical data, such as PCA, are inappropriate. Instead, the analysis should start with distance measures crafted for categorical variables~\cite{Boriah2008}. Following this step, dimensionality reduction methods compatible with distance values, such as Multidimensional Scaling (MDS)~\cite{Leeuw2009}, t-distributed Stochastic Neighbor Embedding (tSNE)~\cite{VanderMaaten2008} and Uniform Manifold Approximation and Projection (UMAP)~\cite{Becht2019}, can be utilized.

\section{Methods}
\label{methods}

\subsection{Overview of TestCat}
\label{results:overview}

\begin{figure}[t]
	\centering
	\includegraphics[width=0.9\linewidth]{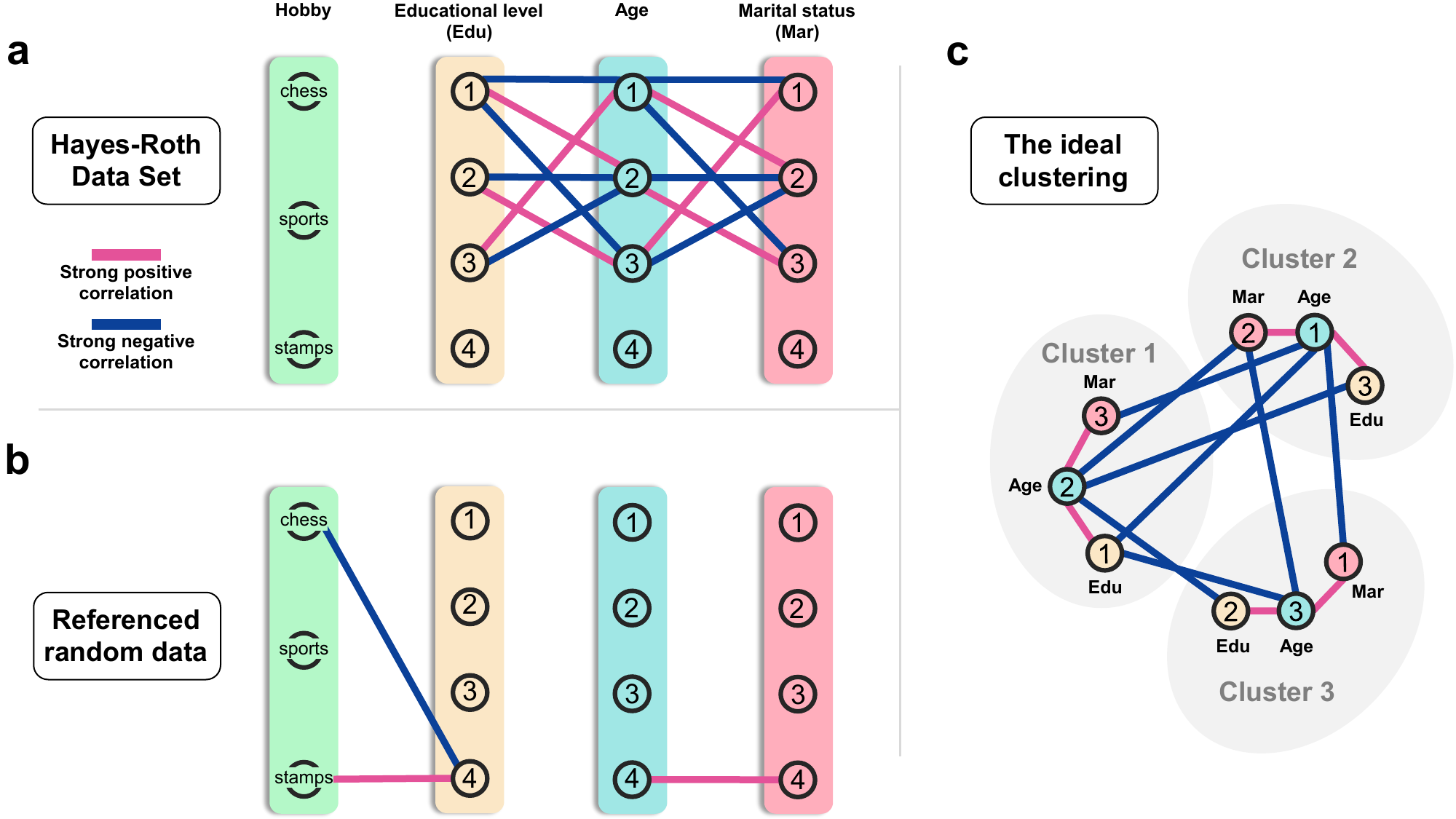}
	\caption{Parallel coordinate plots are utilized to display the strong association among neighboring attribute pairs in both the \textbf{(a)} Hayes-Roth data set and its \textbf{(b)} referenced random data. The attribute values are represented by ``chess", ``sports", ``stamps" or ``1", ``2", ``3", ``4". The strength of association is determined by standardized residuals (refer to Supplementary Method 1). A standardized residual value exceeding $2$ signifies a strong positive association, while a value less than $-2$ indicates a strong negative association. \textbf{(c)} The ideal clustering extracted from Hayes-Roth data set contains attributes with strong positive associations within each cluster, while those with strong negative associations are distributed across separate clusters. Each of these clusters is represented by specific categories.}
	\label{fig:hayes-roth}
\end{figure}

\begin{figure}[t]
	\centering
	\includegraphics[width=1\linewidth]{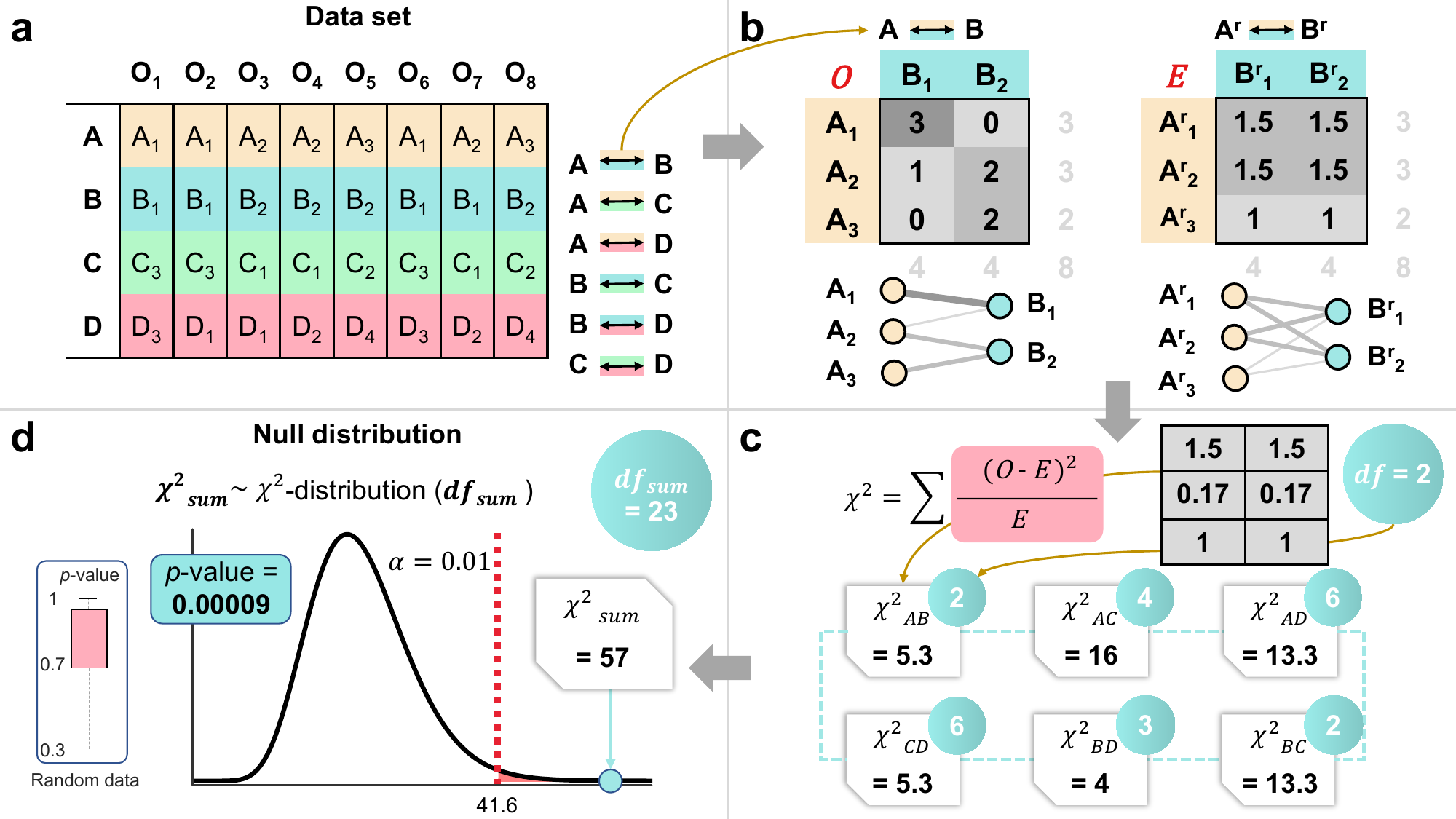}  
	\caption{The TestCat method conducts clusterability analysis on a given categorical data set by providing a $p$-value. \textbf{(a)} A toy categorical data set of four attributes (``A", ``B", ``C", ``D") yields six different attribute pairs that need to be tested. \textbf{(b)} For each attribute pair, such as ``A-B", both its observed ($O$) and expected ($E$) $3\times 2$ contingency tables are constructed and used to calculate a chi-squared test statistic. The darker cells or lines indicate a higher frequency of co-occurrence between the two attribute values in the data set. \textbf{(c)} The chi-squared test statistics for all six attribute pairs, along with their respective degrees of freedom (df), are collected and employed to derive a final $p$-value. \textbf{(d)} Under an imposed assumption (refer to Methods in~\ref{methods:chi}), a single $p$-value can be calculated from the sum of test statistics and its corresponding null distribution. A boxplot of the $p$-values obtained from referenced random data sets is also displayed.} 
	\label{fig:framework}
\end{figure}

TestCat quantifies the clusterability by measuring the strength of association between all pairs of attributes in a given categorical data set. We will demonstrate that a marked difference in association strength exists between clusterable and non-clusterable (random) data. For instance, in the Hayes-Roth data set, exemplars can be grouped into three distinct club memberships, which plays a role in facilitating psychological experiments~\cite{HayesRoth1977}. These qualitative attributes are hobby, educational level, age, and marital status. A conspicuous phenomenon arises from the data (Figure~\ref{fig:hayes-roth}): the attribute values of latter three attributes display an abundance of both positive and negative associations in the original data set (Figure~\ref{fig:hayes-roth}a). Given this strongly associated categories across different attributes, we postulate that the inherent clustering structure (groups) is derived from these distinct attribute combinations as shown in Figure~\ref{fig:hayes-roth}c.  Contrarily, within a random data (generated as explained in Section~\ref{methods:generation}), the values of hobby attribute (a distractor feature) may exhibit a few spurious associated relationship with one attribute value of educational level, while there is only one positive association among the remaining attribute values. We can thus conclude that the original data set is composed of many associated attribute pairs, whereas artificial random data sets lacking a significant clustering structure do not present a comparable magnitude of these associations.

To evaluate the overall strength of association among all attribute pairs and derive a clusterability index in terms of $p$-value, the TestCat approach consists of the following steps (Figure~\ref{fig:framework}). We initiate our analysis by constructing contingency tables for each pair of categorical variables/ attributes (e.g., ``A" and ``B" shown in Figure~\ref{fig:framework}b) of a given data set. Under the assumption of independence between two attributes, the table for the expected frequencies of attribute value pairs can be obtained from (light-colored) marginal frequencies. We employ the chi-squared test to evaluate the association between each pair of attributes. Then, we collect all chi-squared test statistics and their corresponding degrees of freedom for each attribute pair (Figure~\ref{fig:framework}c). Finally, we use the sum of chi-squared test statistics as final test statistic, which will still follows a chi-squared distribution under the assumption that all variables for attribute pairs are independent (Figure~\ref{fig:framework}d). 

The significance testing issue underlying our TestCat method can be formally formulated as follows. Let $S$ denote the number of all attribute pairs. For the $s$-th attribute pair, $H_{0s}$ denotes the null hypothesis that these two attributes are independent and $H_{1s}$ denotes the alternative hypothesis that they are not independent. The global null hypothesis is a composite of the individual null hypotheses $H_{0s}$ for all $S$ tests, represented as:
\begin{equation*}
	H_0: \bigcap_{s=1}^{S} H_{0s},
\end{equation*}
contrasted against the global alternative hypothesis:
\begin{equation}
	H_1: \bigcup_{s=1}^{S} H_{1s},
\end{equation}
where the global alternative hypothesis implies that at least one $H_{0s}$ is false. In the TestCat framework, the global null hypothesis $H_0$ posits that each pair of attributes exhibits no association, where the $s$-th individual null hypothesis $H_{0s}$ aligns with the chi-squared test affirming the independence of the $s$-th attribute pair. Conversely, the global alternative hypothesis $H_1$ indicates the presence of an association between at least one pair of attributes. Furthermore, associations across multiple attribute pairs can cumulatively strengthen $H_1$. To determine whether $H_0$ should be rejected, it is necessary to combine all $p$-values from $S$ individual chi-squared tests into a single composite $p$-value.

Various methods exist for combining $p$-values~\cite{Cinar2022}, such as Fisher's method and Stouffer's method. These approaches create a new test statistic by integrating all $p$-values, which is distinct from the statistic employed in each individual test. In our TestCat approach, we opt for simplicity by using the sum of chi-squared statistics of all attribute pairs as the new test statistic, which also follows a chi-squared distribution under the independence assumption on different pairs of attributes.

\subsection{Chi-squared test}
\label{methods:chi}
In the chi-squared test used in our approach, a chi-squared statistic is obtained for each attribute pair. Consider the $s$-th attribute pair $\langle A,B \rangle_s$, where $A=[A_1,A_2,\cdots,A_{Q^{(s)}_a}]$ and $B=[B_1,B_2,\cdots,B_{Q^{(s)}_b}]$ are two distinct attributes, containing $Q^{(s)}_a$ and $Q^{(s)}_b$ categories respectively. The observed frequencies of attribute value pairs are presented in a contingency table as follows:

\begin{equation}
	O^{(s)}=
	\begin{array}{c|ccc}
		\quad & \text{B}_1 & \cdots & B_{Q^{(s)}_b}  \\
		\hline
		\text{A}_1 & O^{(s)}_{11} & O^{(s)}_{1j} & O^{(s)}_{1Q^{(s)}_b} \\
		\vdots &  O^{(s)}_{i1}  &  O^{(s)}_{ij} & O^{(s)}_{iQ^{(s)}_b} \\
		\text{A}_{Q^{(s)}_a}  & O^{(s)}_{Q^{(s)}_a1} & O^{(s)}_{Q^{(s)}_aj} & O^{(s)}_{Q^{(s)}_aQ^{(s)}_b} \\ 
	\end{array} \eqdot
	\label{eq:OB}
\end{equation}

The $i$-th row and column marginal frequencies, as well as the grand total of the contingency table, are represented as follows:
\begin{equation*}
	O^{(s)}_{i\cdot} = \sum_j^{Q^{(s)}_b}O^{(s)}_{ij}, O^{(s)}_{\cdot j} = \sum_i^{Q^{(s)}_a}O^{(s)}_{ij},
\end{equation*}

\begin{equation}
	O^{(s)}_{\cdot \cdot} = \sum_i^{Q^{(s)}_a}\sum_j^{Q^{(s)}_b}O^{(s)}_{ij} \eqdot
\end{equation}

Then, the expected frequencies can be calculated and represented as follows:
\begin{equation*} 
	E^{(s)}_{ij} = (O^{(s)}_{i\cdot}\cdot O^{(s)}_{\cdot j})/O^{(s)}_{\cdot\cdot},
\end{equation*}
\begin{equation}
	E^{(s)}=
	\begin{array}{c|ccc}
		\quad & \text{B}_1 & \cdots & B_{Q^{(s)}_b}  \\
		\hline
		\text{A}_1 & E^{(s)}_{11}  & E^{(s)}_{1j} & E^{(s)}_{1Q^{(s)}_b} \\
		\vdots &  E^{(s)}_{i1}  &  E^{(s)}_{ij} & E^{(s)}_{iQ^{(s)}_b} \\
		\text{A}_{Q^{(s)}_a} & E^{(s)}_{Q^{(s)}_a1} & E^{(s)}_{Q^{(s)}_aj} & E^{(s)}_{Q^{(s)}_aQ^{(s)}_b} \\
	\end{array} \eqdot
	\label{eq:EX}
\end{equation}

For each attribute pair, we calculate the chi-squared statistic, $\chi_{(s)}^2$, based on $O^{(s)}_{ij}$ and $E^{(s)}_{ij}$. These individual chi-squared statistics are then summed to yield a collective measure for all attribute pairs. The formula is as follows:
\begin{equation}
	\label{eq:sumchi}
	\chi^2_{sum} = \sum_s^S \chi_{(s)}^2
	=  \sum_s^S\sum_i^{Q^{(s)}_a}\sum_j^{Q^{(s)}_b}\frac{(O_{ij}^{(s)}-E_{ij}^{(s)})^2}{E_{ij}^{(s)}},
\end{equation}
where $S=\binom{M}{2}$ denotes the total number of all unique attribute pairs and $M$ is the number of attributes of the given data set. For the $s$-th chi-squared statistic, the degrees of freedom $df^{(s)}$ are equivalent to the number of frequencies in the corresponding contingency table. The aggregate degrees of freedom for the summed chi-squared statistic $\chi_{sum}$, is calculated as follows:
\begin{equation}
	df_{sum} = \sum_s^S df^{(s)} = \sum_s^S(Q^{(s)}_a-1)\cdot(Q^{(s)}_b-1) \eqdot
\end{equation}

Finally, a $p$-value is derived from the summed chi-squared statistic and its aggregate degrees of freedom by utilizing the chi-squared cumulative distribution function. Note that the sum of chi-squared random variables follows a chi-squared distribution if these variables are independent~\cite{Agresti2012}. Here we impose an assumption on the independence among variables for attribute pairs in order to obtain an analytical $p$-value for clusterability evaluation. 

\subsection{Randomized data generation}
\label{methods:generation}
For the $m$-th attribute of randomized data, we assume that both the $m$-th attribute of original data set (ODS) and the $m$-th attribute of its corresponding randomized data set (CRDS) follow the same categorical distribution, which corresponds to the marginal distribution of the contingency table.

To generate a CRDS, we successively generate $M$ random permutations of data objects in ODS. According to the $m$-th permutation, we specify the attribute values for the $m$-th attribute for each data object in CRDS. More precisely, the corresponding attribute value for the $j$-th object in the CRDS is specified to be one that are taken by the $j$-th object of ODS in the permutation. 

\subsection{Theoretical justification}
\label{subsec: thm}
To justify our method, we need to prove that a larger test statistic (corresponding to a smaller $p$-value) indicates the better clusterability of the data set. However, this task is quite difficult in a general setting because (1) the test statistic is complicated since it is the sum of multiple chi-squared statistics, and (2) there is still no consensus on the definition of clusterability of a data set. Nevertheless, we try to provide a theoretical justification on a special case that the data set is only composed of two categorical attributes. More precisely, we first investigate the case that each attribute is composed of only two distinct attribute values. Then, we extend the analysis to the general case that each attribute can have more than two attribute values.

\subsubsection{Special case: binary categorical attributes}

Suppose that these two attributes are denoted by $A=\{A_1,A_2\}$ and $B=\{B_1,B_2\}$, respectively. The corresponding contingency table is given as follows, where $C_{ij}$ is the number of co-occurrence of attribute values $A_{i}$ and $B_{j}$. Obviously, such a data set is composed of four natural clusters, where each cluster is characterized  by $A=A_{i}$ and $B=B_{j}$. In other words, each cluster consists of $C_{ij}$ identical attribute value pairs. Under this setting, it is easy to see that the distances among samples within each cluster are all zeros. Hence, to quantify the clusterability of such a data set, we only need to check the distances between samples from different clusters. More precisely, the following measure is employed to quantify the clusterability of a data set.

\begin{equation}
	\begin{array}{|c|cc|c}
		\cline{2-3}
		\multicolumn{1}{c|}{} & \text{B}_1 & \text{B}_2 & \\
		\cline{1-3}
		\ \text{A}_1 \ \ & C_{11} & C_{12} & \ C_{1\cdot} \\
		\ \text{A}_2 \ \ & C_{21} & C_{22} & \ C_{2\cdot}\\ 
		\cline{1-3}
		\multicolumn{1}{c}{} & \multicolumn{1}{c}{C_{\cdot 1}} & \multicolumn{1}{c}{C_{\cdot 2}} & \ C_{\cdot\cdot} \\
	\end{array} 
	\label{eq:tab}
\end{equation}

\begin{definition}
	Let the data set $DS$ be characterized by the contingency table shown in~\ref{eq:tab}. The normalized separation of $DS$, denoted as $Sep_{norm}(DS)$, is defined as follows:
	
	\begin{equation}
		Sep_{norm}(DS) = \frac{Sep(DS)}{S_{total}(DS)},
	\end{equation}
	where $Sep(DS)$ is the sum of all pairwise inter-cluster Hamming distances and $S_{total}$ is the number of all pairwise inter-cluster samples. The concise expression for $Sep(DS)$ can be calculated as follows:
	\begin{equation}
		\begin{split}
			Sep(DS) & = Sep(C_{11},C_{12})+ Sep(C_{11},C_{21})+ Sep(C_{11},C_{22})\\ & + Sep(C_{12},C_{21})+ Sep(C_{12},C_{22})+ Sep(C_{21},C_{22})  \\ & =
			C_{11}C_{12} + C_{11}C_{21} + 2C_{11}C_{22} + 2C_{12}C_{21} + C_{12}C_{22} + C_{21}C_{22}\\  & =
			C_{11}(C_{\cdot\cdot}-C_{11}) + C_{22}(C_{\cdot\cdot}-C_{22}) + 2C_{12}C_{21}
		\end{split}\eqdot
	\end{equation}
	Given $C_{22}=C_{\cdot 2}-(C_{1\cdot}-C_{11}) \text{ and } C_{12}C_{21}=(C_{1\cdot}-C_{11})(C_{\cdot 1}-C_{11})$, we have $Sep(DS) = C_{11}(C_{\cdot\cdot}-C_{11}) + (C_{11}-C_{1\cdot}+C_{\cdot 2})(C_{\cdot\cdot}-C_{11}+C_{1\cdot}-C_{\cdot 2})\\ + 2(C_{1\cdot}-C_{11})(C_{\cdot 1}-C_{11})$. If we set $C_{11}=n, C_{\cdot\cdot}=N, C_{1\cdot}=x, C_{\cdot 1}=y, C_{\cdot 2}=z$, then we can write as:
	\begin{equation*}
		\begin{split}
			Sep(DS) & = n(N-n) + (n-x+z)(N-n+x-z) + 2(x-n)(y-n)\\ & = 
			Nn - n^2 + (N-n+x-z)n - x(N-n+x-z) \\ & + z(N-n+x-z) + 2xy + 2n^2 -2(x+y)n\\ & =
			(2N-2z-2y)n+(z-x)N-x^2-z^2  +2x(y+z)
		\end{split}\eqdot
	\end{equation*}
	Given $y+z=C_{\cdot 1}+C_{\cdot 2}=C_{\cdot\cdot}=N$, we have
	\begin{equation}
		\begin{split}
			Sep(DS) & = (z-x)N - (x^2+z^2) + 2xN \\ & = (x+z)N-((x+z)^2-2xz) \\ & = (N-x-z)(x+z)+2xz \\ & = 
			(C_{\cdot\cdot}-C_{1\cdot}-C_{\cdot 2})(C_{1\cdot}+C_{\cdot 2})+2C_{1\cdot}C_{\cdot 2}
		\end{split}\eqdot
	\end{equation}
	
	Now we turn to express another element $S_{total}(DS)$ as follows:
	\begin{equation}
		\begin{split}
			S_{total}(DS) & = S(C_{11},C_{12})+ S(C_{11},C_{21})+ S(C_{11},C_{22})\\ & + S(C_{12},C_{21})+ S(C_{12},C_{22})+ S(C_{21},C_{22})  \\ & =
			C_{11}C_{12}+C_{11}C_{21}+C_{11}C_{22} +C_{12}C_{21}+C_{12}C_{22} + C_{21}C_{22} \\ & = 
			\big(C_{11}(C_{\cdot\cdot}-C_{11}) + C_{22}(C_{\cdot\cdot}-C_{22}) + 2C_{12}C_{21}\big) \\ & - (C_{12}C_{21}+C_{11}C_{22}) \\ & = 
			Sep(DS) - (C_{12}C_{21}+C_{11}C_{22})
		\end{split}\eqdot
	\end{equation}
\end{definition}

\begin{lemma} 
	Given a $2\times 2$ contingency table with fixed marginal frequencies, if $C_{11}$ exceeds $\lambda=\frac{2C_{1\cdot}+C_{\cdot 1}-C_{\cdot 2}}{4}$, then the $Sep_{norm}(DS)$ is directly proportional to $C_{11}$: $C_{11}>\lambda \to Sep_{norm}(DS)\propto C_{11}$.
\end{lemma}
\begin{proof}
	Given that $Sep_{norm}(DS) = \frac{Sep(DS)}{Sep(DS) - (C_{12}C_{21} + C_{11}C_{22})}$ where $Sep(DS)$ is a constant under fixed marginal frequencies, $Sep_{norm}(DS)$ is only dependent on the term $(C_{12}C_{21} + C_{11}C_{22})$. Thus we rewrite $(C_{12}C_{21} + C_{11}C_{22})$ with $C_{11}$ as the only variable as follows:
	\begin{equation}
		\begin{split}
			C_{12}C_{21} + C_{11}C_{22} & = C_{11}(C_{\cdot 2}-(C_{1\cdot}-C_{11}))  \\ & +(C_{1\cdot}-C_{11})(C_{\cdot 1}-C_{11}) \\ & = 
			C_{\cdot 2}C_{11}-C_{1\cdot}C_{11}+(C_{11})^2  \\ & +
			C_{1\cdot}C_{\cdot 1}-C_{1\cdot}C_{11}-C_{\cdot 1}C_{11}+(C_{11})^2 \\ & =
			2(C_{11})^2 + (C_{\cdot 2}-2C_{1\cdot}-C_{\cdot 1})C_{11} + C_{1\cdot}C_{\cdot 1}\\ & =
			f(C_{11})
		\end{split}\eqdot
	\end{equation}
	
	Taking the derivative of $f$ with respect to $C_{11}$, we get
	$f'(C_{11}) = 4C_{11} + (C_{\cdot 2} - 2C_{1\cdot} - C_{\cdot 1})$ and find the critical point: $\lambda = \frac{2C_{1\cdot}+C_{\cdot 1}-C_{\cdot 2}}{4}$.
	
	Thus, when $C_{11} > \lambda$, $f(C_{11})$ is an increasing function. Now, since $Sep_{norm}(DS)$ is functionally dependent on $f(C_{11})$, it follows that the clusters become more distinct in terms of separation
	as $C_{11}$ increases, thereby supporting the claim that a larger $C_{11}$ within the $2\times 2$ contingency table corresponds to more distinct clusters.
\end{proof}

\begin{lemma}
	Given a $2 \times 2$ contingency table with fixed marginal frequencies, assume without loss of generality that $C_{11}$ is greater than its expected frequency $E_{11}$. Under this condition, the chi-squared statistic $\chi^2(DS)$ is directly proportional to $C_{11}$: if $C_{11} > E_{11}$, then $\chi^2(DS) \propto C_{11}$.
\end{lemma}

\begin{proof}
	Given the condition $C_{11}>E_{11}=\frac{C_{1\cdot}C_{\cdot 1}}{C_{\cdot\cdot}}$, we have $C_{\cdot\cdot}C_{11} - C_{1\cdot}C_{\cdot 1} > 0$. Furthermore, according to the definition of the chi-squared statistic for a $2\times 2$ contingency table, we have  $\chi^2(DS) = \frac{C_{\cdot \cdot}(C_{11}C_{22}-C_{12}C_{21})^2}{C_{1\cdot}C_{\cdot 1}C_{\cdot 2}C_{2\cdot}} =  C_{DS}(C_{11}C_{22}-C_{12}C_{21})^2$ where $C_{DS}$ is a constant under fixed marginal frequencies. The $\chi^2(DS)$ is only dependent on the squared term $(C_{11}C_{22}-C_{12}C_{21})^2$. Thus we can express $(C_{11}C_{22}-C_{12}C_{21})$ with $C_{11}$ as the only variable as follows:
	\begin{equation}
		\label{eq:lm2}
		\begin{split}
			C_{11}C_{22}-C_{12}C_{21} & = 
			C_{11}(C_{\cdot 2}-(C_{1\cdot}-C_{11})) -(C_{1\cdot}-C_{11})(C_{\cdot 1}-C_{11}) \\ & =
			C_{\cdot 2}C_{11}-C_{1\cdot}C_{11}+(C_{11})^2 \\ & - 
			C_{1\cdot}C_{\cdot 1}+C_{1\cdot}C_{11}+C_{\cdot 1}C_{11}-(C_{11})^2 \\ & =
			(C_{\cdot 2}-C_{1\cdot}+C_{1\cdot}+C_{\cdot 1})C_{11} - C_{1\cdot}C_{\cdot 1} \\ & =
			C_{\cdot\cdot}C_{11} - C_{1\cdot}C_{\cdot 1} \\ &  > 0
		\end{split}\eqdot
	\end{equation}
	
	Now that we have established $C_{11}C_{22}-C_{12}C_{21}=C_{\cdot\cdot}C_{11} - C_{1\cdot}C_{\cdot 1}$ according to Equation~\ref{eq:lm2}. Under the given condition $C_{11}>E_{11}$, $(C_{\cdot\cdot}C_{11} - C_{1\cdot}C_{\cdot 1})^2=(C_{11}C_{22}-C_{12}C_{21})^2$ is an increasing function. This indicates a stronger association between attributes A and B as $C_{11}$ increases, thereby supporting the claim that a larger $C_{11}$ within the $2\times 2$ contingency table is indicative of an increased strength of association.
\end{proof}

\begin{theorem}
	\label{thm}
	Given a $2 \times 2$ contingency table with fixed marginal frequencies, for any two data sets $DS_1$ and $DS_2$ sampled according to these marginal distributions, if $\chi^2(DS_1) > \chi^2(DS_2) > \lambda^*=C_{DS}\big(C_{\cdot\cdot}\lambda - C_{1\cdot}C_{\cdot 1}\big)^2$ where $C_{DS}=\frac{C_{\cdot \cdot}}{C_{1\cdot}C_{\cdot 1}C_{\cdot 2}C_{2\cdot}}$ and $\lambda = \frac{2C_{1\cdot}+C_{\cdot 1}-C_{\cdot 2}}{4}$, then it follows that $Sep_{norm}(DS_1) > Sep_{norm}(DS_2)$. 
\end{theorem}

\begin{proof}
	Let us denote $p$ and $q$ as the count variables in the $2 \times 2$ contingency table corresponding to the positively associated cell $C_{11}$ for data sets $DS_1$ and $DS_2$, respectively. Since the marginal frequencies are fixed, the chi-squared statistics $\chi^2(DS_1)$ and $\chi^2(DS_2)$ along with the same constant can be expressed in terms of $p, q$ as $C_{DS}(C_{\cdot\cdot}p - C_{1\cdot}C_{\cdot 1})^2$  and $C_{DS}(C_{\cdot\cdot}q - C_{1\cdot}C_{\cdot 1})^2$ according to Equation~\ref{eq:lm2}. Then we can rewrite the condition $\chi^2(DS_2)>\lambda^*$ as follows:
	
	\begin{equation}
		\label{eq:thm}
		\begin{split}
			\chi^2(DS_2)>\lambda^* & \Rightarrow 
			C_{DS}(C_{\cdot\cdot}q - C_{1\cdot}C_{\cdot 1})^2 \\ & > C_{DS}(C_{\cdot\cdot}\lambda - C_{1\cdot}C_{\cdot 1})^2 \\ & \Rightarrow 
			C_{\cdot\cdot}q - C_{1\cdot}C_{\cdot 1} \\ & > C_{\cdot\cdot}\lambda - C_{1\cdot}C_{\cdot 1}\\ & \Rightarrow  
			q > \lambda = \frac{2C_{1\cdot}+C_{\cdot 1}-C_{\cdot 2}}{4}
		\end{split}\eqdot
	\end{equation}
	
	Combining the results from Lemmas 1 and 2 with Equation~\ref{eq:thm}, we finalize the proof as follows: Lemma 2 provides that $\chi^2(DS)$ is directly proportional to $C_{11}$. Consequently, given $\chi^2(DS_1) > \chi^2(DS_2) > \lambda^*$, we apply Equation~\ref{eq:thm} to deduce that $p > q > \lambda$. Furthermore, Lemma 1 suggests that for $C_{11} > \lambda$, which is satisfied here as $p > q > \lambda$, the normalized separation measure $Sep_{norm}(DS)$ will increase. Therefore, we can assert that $Sep_{norm}(DS_1)$ is greater than $Sep_{norm}(DS_2)$.
	
\end{proof}

\subsubsection{Extension to general categorical attributes}

We now extend our theoretical justification to the case where the two attributes $A$ and $B$ are categorical variables with more than two categories. Let $A = \{A_1, A_2, \cdots, A_r\}$ and $B = \{B_1, B_2, \dotsc, B_s\}$, where $r$ and $s$ are the numbers of distinct categories of $A$ and $B$, respectively.

An important observation is that the Hamming distance between two samples computed on the original categorical variables is identical to the Hamming distance computed after one-hot encoding these variables. Given this equivalence, we can conceptually extend our previous results from the binary case to the general case. The chi-squared statistic for the contingency table of $A$ and $B$ is:

\begin{equation}
	\chi^2(DS) = \sum_{i=1}^{r} \sum_{j=1}^{s} \frac{\left( C_{ij} - E_{ij} \right)^2}{E_{ij}},
\end{equation}
where $C_{ij}$ is the observed frequency of the category pair $(A_i, B_j)$, and $E_{ij} = \frac{C_{i\cdot} C_{\cdot j}}{N}$ is the expected frequency under independence.

A larger chi-squared statistic indicates stronger associations between certain categories of $A$ and $B$, which promotes the formation of clusters with higher intra-cluster similarity and greater inter-cluster separation. Since each summation term in the equation above can be interpreted as the case of binary categorical attributes, we can infer that this cumulative larger chi-squared statistic corresponds to a higher normalized separation measure $Sep_{norm}(DS)$, indicating better clusterability of the data set.

This extension suggests that our method is applicable to attributes with more than two categories, and that the relationship between the $\chi^2(DS)$ and the $Sep_{norm}(DS)$ holds in the general case. However, we acknowledge that deriving an explicit analytical form in this general setting is complex and challenging, requiring more in-depth analysis and effort.

\section{Results}
\label{results}
\subsection{An illustrative example for two attributes}
In the context of a $2\times2$ contingency table, under certain mild conditions, we have demonstrated in Theorem~\ref{thm} that one data set with a larger chi-squared test statistic will exhibit a better clusterability. As an illustrative example, let us consider a series of data sets, each containing 100 students, with two attributes: ``grades in math'' and ``grades in physics'', where the co-occurrence of categories ``Good-in-Math'' and ``Good-in-Physics'' indicates a positive association, represented by the count $C_{11}$, consistent with the notation used in Section~\ref{subsec: thm}.

We set $C_{11}$ to be 20, 15, and 10 on three different data sets, each with fixed marginal distributions ($C_{\cdot 1},C_{\cdot 2},C_{1\cdot},C_{2\cdot}$). These data sets are designated as $DS_1,DS_2,DS_3$ respectively. The contingency tables for these data sets are as follows:

\begin{center}
	\begin{tabular}{|c|c|c|c}
		\cline{1-3}
		$DS_1$ & Good-in-Physics & Poor-in-Physics & \\
		\cline{1-3}
		Good-in-Math & 20 (Cluster 1)&  5  (Cluster 2) & 25 \\
		\cline{1-3}
		Poor-in-Math & 20 (Cluster 3)&  55 (Cluster 4)& 75 \\
		\cline{1-3}
		\multicolumn{1}{c}{} & \multicolumn{1}{c}{40} & \multicolumn{1}{c}{60} & \multicolumn{1}{c}{100}
	\end{tabular}
\end{center}
\begin{center}
	\begin{tabular}{|c|c|c|c}
		\cline{1-3}
		$DS_2$ & Good-in-Physics & Poor-in-Physics & \\
		\cline{1-3}
		Good-in-Math & 15 (Cluster 1)&  10  (Cluster 2) & 25 \\
		\cline{1-3}
		Poor-in-Math & 25 (Cluster 3)&  50 (Cluster 4)& 75 \\
		\cline{1-3}
		\multicolumn{1}{c}{} & \multicolumn{1}{c}{40} & \multicolumn{1}{c}{60} & \multicolumn{1}{c}{100}
	\end{tabular}
\end{center}
\begin{center}
	\begin{tabular}{|c|c|c|c}
		\cline{1-3}
		$DS_3$ & Good-in-Physics & Poor-in-Physics & \\
		\cline{1-3}
		Good-in-Math & 10 (Cluster 1) &  15  (Cluster 2) & 25 \\
		\cline{1-3}
		Poor-in-Math & 30 (Cluster 3) &  45 (Cluster 4) & 75 \\
		\cline{1-3}
		\multicolumn{1}{c}{} & \multicolumn{1}{c}{40} & \multicolumn{1}{c}{60} & \multicolumn{1}{c}{100} 
	\end{tabular}
\end{center}

From the above contingency tables, we can calculate their corresponding $\chi^2$, $p$-value and $Sep_{norm}$ as follows:
\begin{center}
	\begin{tabular}{|c|c|c|c|}
		\hline
		Measure & $DS_1$ & $DS_2$ & $DS_3$ \\
		\hline
		$\chi^2$ & 22.2 & 5.6 & 0 \\
		\hline
		$p$-value & 2.4E-06 &  0.02 & 1 \\
		\hline
		$Sep_{norm}$ & 1.39 & 1.31 & 1.27 \\
		\hline 
	\end{tabular}
\end{center}
Consistent with Theorem~\ref{thm}, if $\chi^2(DS_1) > \chi^2(DS_2) > \chi^2(DS_3)$, it follows that more distinct clusters can be formed in $DS_1$ than in $DS_2$ and $DS_3$, as demonstrated by the inequality $Sep_{norm}(DS_1) > Sep_{norm}(DS_2) > Sep_{norm}(DS_3)$. Notably, $DS_3$ represents completely randomized data since each cell count equals its expected value, akin to a single, indivisible cluster from a Gaussian distribution in numerical data, which suggests an absence of clustering structure. Furthermore, our method provides a $p$-value, inherently establishing a statistically sound threshold (e.g., $p$-value$\leq 0.01$), at which our method can determine that only $DS_1$ is clusterable. In $DS_1$, 75 students can be perfectly grouped into two clusters: ``Good-in-Physics \& Good-in-Math'' and ``Poor-in-Physics \& Poor-in-Math''. However, in $DS_2$ and $DS_3$, with only 65 and 55 students respectively able to be grouped, all grades are not considered to have sufficient association.

\begin{table}[b]
	\small
	\caption{The properties of 18 UCI data sets.}
	\label{tab:UCI}
	\centering
	\begin{tabular}{@{}ccccc@{}} 
			\hline
			Data set & \textit{Abbr.}  & \#Objects & \#Attributes & \#Categories \\
			\hline
			Lenses & Ls & 24 & 4 & 9 \\
			Lung Cancer & Lc  & 32 & 56 & 159 \\
			Soybean (Small) & So  & 47 & 21 & 58  \\
			Zoo & Zo  & 101 & 16 & 36 \\
			Promoter Sequences & Ps  & 106 & 57 & 228 \\
			Hayes-Roth & Hr  & 132 & 4 & 15 \\
			Lymphography & Ly  & 148 & 18 & 59 \\
			Heart Disease & Hd  & 303 & 13 & 57 \\
			Solar Flare & Sf  & 323 & 9 & 25 \\
			Primary Tumor & Pt  & 339 & 17 & 42 \\
			Dermatology & De  & 366 & 33 & 129 \\
			House Votes & Hv  & 435 & 16 & 48 \\
			Balance Scale & Bs & 625 & 4 & 20 \\
			Credit Approval & Ca  & 690 & 9 & 45 \\
			Breast Cancer & Bc  & 699 & 9 & 90 \\
			Mammographic Mass & Mm  & 824 & 4 & 18 \\
			Tic-Tac-Toe & Tt  & 958 & 9 & 27 \\
			Car Evaluation & Ce  & 1728 & 6 & 21 \\
			\hline
	\end{tabular}
\end{table}

\subsection{Data sets and validation strategies}
To illustrate and evaluate TestCat, we initiate our study with a selection of commonly used categorical data sets from the UCI repository~\cite{Dua2019}, many of which are likely to have natural clustering structure drawn from various subject areas including life, social, physical, financial fields. The properties of 18 UCI data sets utilized in our study are outlined in Table~\ref{tab:UCI}. 

Since there are still no recognized benchmark data sets and validation methodology for evaluating the clusterability of categorical data sets, we adopt the following strategy in the performance comparison. Our approach presupposes that all data sets from UCI are clusterable, while their corresponding randomized data sets lack a clustering structure. The details of our validation strategy are elaborated below. 

For each original data set (ODS), we generate its corresponding randomized data set (CRDS) (generated as explained in Section~\ref{methods:generation}). If a data set has a clustering structure, an effective clusterability evaluation method should be able to recognize its ODS as being clusterable while identifying its CRDS as being unclusterable. That is, each ODS is a true positive (clusterable) and all their CRDSs are false positives (unclusterable). Conversely, for a data set devoid of a clustering structure, it is desirable to correctly identify both its ODS and CRDS as being unclusterable. In all scenarios, the ability to discern random data constitutes an essential requirement for any trustworthy clusterability evaluation method.

Since CRDS is not unique, we generate a representative CRDS from a pool of multiple CRDSs for each ODS: Initially, we generate 101 randomized data sets for each ODS. The distribution of $p$-values (calculated by TestCat) of these 101 randomized data sets is shown in Supplementary Figure 1. In most cases, the $p$-values of these 101 CRDSs exceed 0.01, as displayed in Supplementary Table 1. From these pool, we obtain a median $p$-value. Subsequently, we generate additional CRDSs. The chosen CRDS is the first one sequentially generated over multiple runs, with its $p$-value deviating no more than 0.05 from the median.

\begin{figure}[t]
	\centering
	\includegraphics[width=1\linewidth]{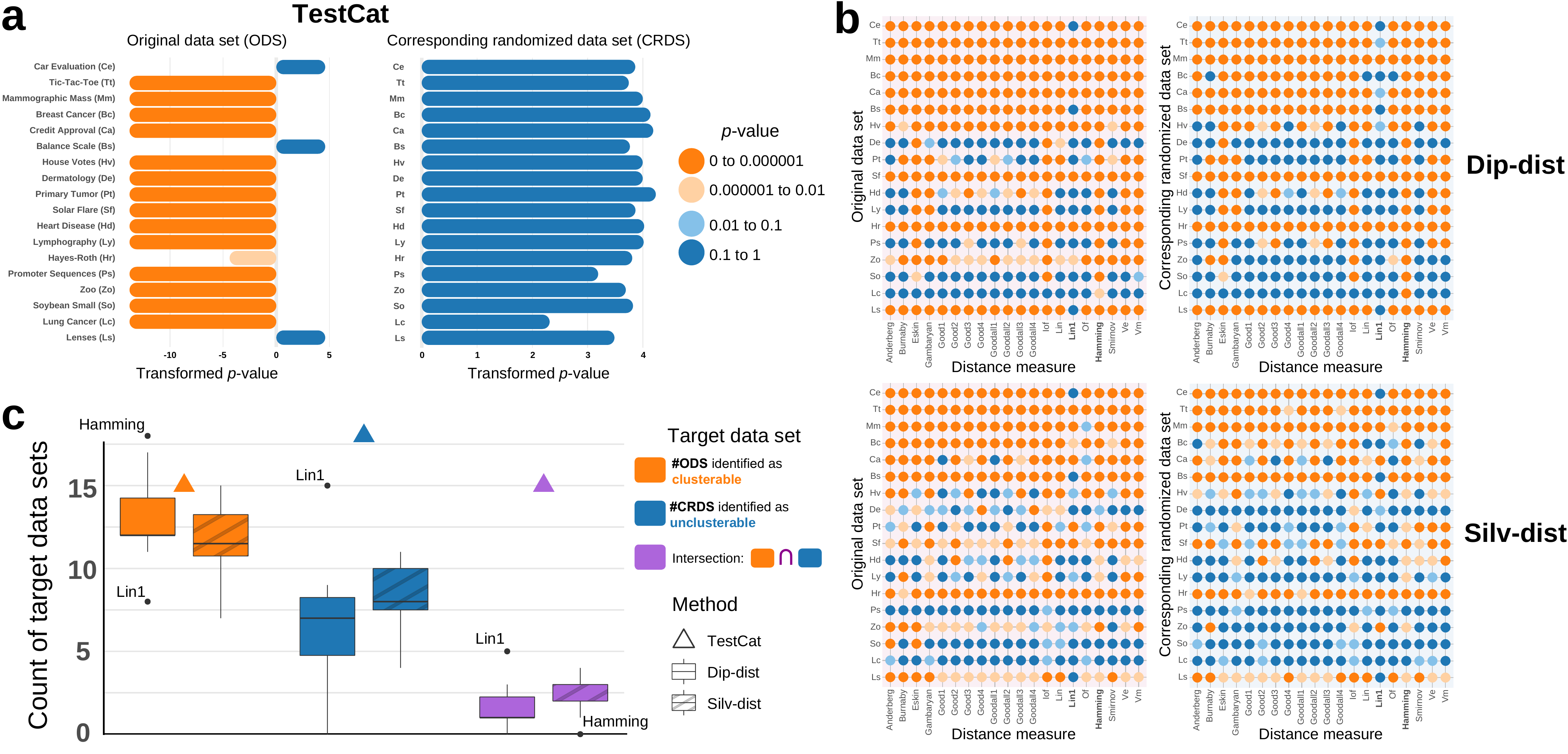} 
	\caption{Identification results of TestCat and existing methods without dimensionality reduction on 18 UCI data sets (including original data sets and corresponding randomized data sets). \textbf{(a)} The barplots of $p$-values produced by TestCat. To better visualize smaller $p$-values that approach or equal 0, and to distinguish the significance level of 0.01 through the y-axis ($y=0$), we use transformed $p$-values, defined by the formula: $y=\log (p\text{-value}/ 0.01 + 0.000001)-\log (0.000001)$. \textbf{(b)} Dip and Silverman have been applied to distance values obtained from 20 different categorical distance measures (detailed descriptions of all these measures can be found in reference~\cite{Sulc2022}), hence constituting two sets of comparison methods: Dip-dist and Silv-dist. The outcomes of these comparisons are illustrated in heatmap. \textbf{(c)} We evaluate TestCat against Dip-dist and Silv-dist by counting the number of correctly identified ODSs and CRDSs under the significance level of 0.01. The boxplots describe the count distributions of variants of Dip-dist and Silv-dist derived from the 20 distance measures employed. Outliers are denoted in accordance with the specific distance measure used.} 
	\label{fig:Metrics-cate}
\end{figure}

\subsection{Compared methods}
Since there are still no clusterability evaluation methods that are specially developed for categorical data, we adopt and repurpose methodologies typically applicable to numerical data~\cite{Adolfsson2019}. These methods consist of two key elements: obtaining one-dimensional transformed values and conducting multimodality tests. Dip test (abbreviated as Dip) and Silverman test (abbreviated as Silv) are among the commonly used multimodality tests~\cite{Adolfsson2019,Laborde2023}. To obtain the one-dimensional representation, one method is to acquire pairwise distance through measures specifically crafted for categorical data, and the other is to employ similarity-preserved dimensionality reduction. In our experiment, we trialed 20 different categorical distance measures~\cite{Boriah2008}, and subsequently applied Dip and Silv tests, denoted as Dip-dist and Silv-dist, respectively. When deploying similarity-preserved dimensionality reduction techniques, we utilized MDS~\cite{Leeuw2009}, tSNE~\cite{VanderMaaten2008} and UMAP~\cite{Becht2019}, denoted as MDS Dip/ Silv, tSNE Dip/ Silv, and UMAP Dip/ Silv, respectively. 

Here, we clarify the inherent stochastic factors in some of these dimensionality reduction methods and their implications on our experimental setup. For MDS, we leverage a version that incorporates a common heuristic strategy to select the solution with the lowest stress value from several configurations. The initial configuration is set randomly. Analogously, tSNE and UMAP involve randomness in their optimization processes and the initialization of embedding. Therefore, these techniques might converge to different local optimal solutions, resulting in varied outputs across multiple runs. On the contrary, PCA and SPCA are deterministic algorithms that invariably generate consistent outputs for a given set of input data. Given these characteristics, in our study, we execute MDS, tSNE, and UMAP multiple times to account for their intrinsic stochasticity. For PCA and SPCA, a single run is sufficient due to their deterministic nature.

In addition to the aforementioned comparative methods, certain advanced embedding methods such as CDE and CDC\_DR~\cite{Bai2022} are designed to transform categorical data into numerical data for clustering tasks. This transformation enables the direct application of conventional clusterability evaluation techniques, which are typically suitable for numerical data. However, these Categorical-to-Numerical methods tend to generate high-dimensional numerical representations. To tackle this issue in practice, as suggested in reference~\cite{Laborde2023}, we employ dimensionality reduction tools such as PCA and Sparse Principal Component Analysis (SPCA). These techniques allow us to condense the high-dimensional data into manageable one-dimensional values.  We then apply Dip and Silv on that, referred to as PCA Dip/ Silv and SPCA Dip/ Silv, respectively.

\subsection{Performance comparison results}
\label{section:comparison}

\begin{table}[b]
	\small
	\caption{The clusterability analysis results of TestCat on 18 UCI data sets. ``\#Pairs'' indicates the number of pairs of attribute values exhibiting strong positive or negative association, with the percentage shown in parentheses representing the proportion of such associated pairs. The distribution of these \#Pairs from CRDSs (101 runs) for each ODS is provided in Supplementary Figure 1b.} 
	\label{tab:resultsTestCat}
	\begin{tabular}{|c|cc|cc|}
		\hline
		&  \multicolumn{2}{c|}{ODS} &  \multicolumn{2}{c|}{CRDS}  \\
		\multicolumn{1}{|c|}{\multirow{-2}{*}{\centering Data set}} & \#Pairs & $p$-value & \#Pairs  & $p$-value \\
		\hline
		\textbf{Ls} & 0 (0\%) & 1 & 2 (6.7\%)  & \textcolor{db}{\textbf{0.32}} \\
		\textbf{Lc}  & 1167 (9.4\%) & \textcolor{dr}{\textbf{5E-111}} & 570 (4.6\%) & \textcolor{db}{\textbf{0.10}} \\
		\textbf{So}  & 472 (29.8\%) & \textcolor{dr}{\textbf{7E-248}} & 62 (3.9\%) & \textcolor{db}{\textbf{0.45}} \\
		\textbf{Zo}  & 306 (51.0\%) & \textcolor{dr}{\textbf{2E-267}} & 26 (4.3\%) & \textcolor{db}{\textbf{0.40}} \\
		\textbf{Ps}  & 1537 (6.0\%) & \textcolor{dr}{\textbf{5E-26}} & 1141 (4.5\%) & \textcolor{db}{\textbf{0.24}} \\
		\textbf{Hr}  & 21 (25.0\%) & \textcolor{lr}{\textbf{1E-4}} & 2 (2.4\%) & \textcolor{db}{\textbf{0.45}} \\
		\textbf{Ly}  & 358 (22.2\%) & \textcolor{dr}{\textbf{8E-195}} & 82 (5.1\%) & \textcolor{db}{\textbf{0.55}} \\
		\textbf{Hd}  & 250 (17.0\%) & \textcolor{dr}{\textbf{1E-79}}  & 63 (4.3\%) & \textcolor{db}{\textbf{0.55}} \\
		\textbf{Sf}  & 133 (49.2\%) & \textcolor{dr}{\textbf{5E-180}}  & 15 (5.6\%) & \textcolor{db}{\textbf{0.47}} \\
		\textbf{Pt}  & 250 (30.3\%) & \textcolor{dr}{\textbf{1E-101}}  & 56 (6.8\%) & \textcolor{db}{\textbf{0.68}} \\
		\textbf{De}  & 2719 (33.7\%) & \textcolor{dr}{\textbf{0}}  & 330 (4.1\%) & \textcolor{db}{\textbf{0.54}} \\
		\textbf{Hv}  & 585 (54.2\%) & \textcolor{dr}{\textbf{0}}  & 55 (5.1\%) & \textcolor{db}{\textbf{0.54}} \\
		\textbf{Bs} & 0 (0\%) & 1 & 13 (8.7\%) & \textcolor{db}{\textbf{0.43}} \\
		\textbf{Ca}  & 219 (26.7\%) & \textcolor{dr}{\textbf{0}}  & 36 (4.4\%) & \textcolor{db}{\textbf{0.65}} \\
		\textbf{Bc}  & 1296 (36.0\%) & \textcolor{dr}{\textbf{0}}  & 156 (4.3\%) & \textcolor{db}{\textbf{0.62}} \\
		\textbf{Mm}  & 44 (36.4\%) & \textcolor{dr}{\textbf{2E-194}} & 5 (4.1\%) & \textcolor{db}{\textbf{0.54}} \\
		\textbf{Tt}  & 106 (32.7\%) & \textcolor{dr}{\textbf{4E-106}}  & 18 (5.6\%) & \textcolor{db}{\textbf{0.42}} \\
		\textbf{Ce}  & 0 (0\%) & 1 & 10 (5.5\%) & \textcolor{db}{\textbf{0.47}} \\
		\hline
	\end{tabular}
\end{table}

\begin{figure}[t]
	\centering
	\includegraphics[width=1\linewidth]{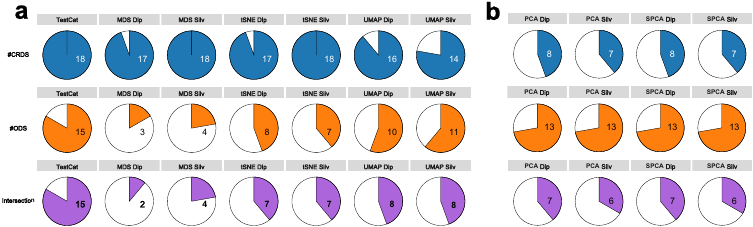} 
	\caption{Count of correctly identified data sets by using TestCat and compared methods. \textbf{(a)} Compared methods via dimensionality reduction (running 101 times for each data set) based on \textbf{Hamming} distance. The resulting median $p$-value from 101 runs is used for determining whether each target data set is clusterable. The experimental results based on Lin1 distance are displayed in Supplementary Figure 2. \textbf{(b)} Categorical-to-Numerical methods via CDC\_DR embedding. We implemented the CDC\_DR embedding on each categorical data set to generate its numerical representation. Following this transformation, we utilized PCA or SPCA to further condense this numerical data. The experimental results based on CDE embedding are displayed in Supplementary Figure 3.} 
	\label{fig:Count-Hamming}
\end{figure}

We now turn to the evaluation and comparison of TestCat with other clusterability evaluation methods, employing the validation strategy described above. Our investigation reveals that, using TestCat for evaluation, most of these UCI data sets are identified as being clusterable and all their CRDSs are identified as being unclusterable (see Figure~\ref{fig:Metrics-cate}a and Table~\ref{tab:resultsTestCat}), a conclusion not necessarily guaranteed by existing methods (see Figure~\ref{fig:Metrics-cate}c, Figure~\ref{fig:Count-Hamming}, Supplementary Figure 2 and Supplementary Figure 3). More details are as follows.

Initially, we examine existing methods that circumvent dimensionality reduction. As depicted in Figure~\ref{fig:Metrics-cate}b, none of these methods manage to identify all CRDSs as being unclusterable. Among twenty measures evaluated, only six successfully identify more than half of all target CRDSs ($\text{\#CRDS}\geq 10$): both Dip-dist and Silv-dist methods based on Lin1, along with Silv-dist method based on Good1, Good4, Goddall1, Goddall4, and Of. When it comes to identifying more than half of all ODSs as being clusterable ($\text{\#ODS}\geq 10$), most methods accomplish this feat, except for Dip-dist using Lin1 and Silv-dist using Good1, Goddall1, and Of. Intriguingly, Dip-dist method utilizing Hamming distance correctly identifies every ODS but remarkably fails to recognize any CRDSs. As demonstrated in Figure~\ref{fig:Metrics-cate}c), no single method is able to simultaneously and correctly recognize all ODSs and CRDSs. More precisely, we can employ the count of data sets in the intersection of correctly identified ODS and CRDS as a quantitative indicator. Even the most effective existing method, Dip-dist based on Lin1, could only identify an intersection set with five data sets: Zo, Hv, Ca, Bc, and Tt.

In light of above experimental results, Hamming and Lin1 distances display distinguished performance with respect to at least one or two of the three count metrics (as represented by the outliers in Figure~\ref{fig:Metrics-cate}c): number of correctly identified ODS, number of correctly identified  CRDS, and the size of intersection set. Nevertheless, all these methods cannot correctly identify all CRDSs, thus failing to meet the trustworthy standard required to discern random data. Consequently, we leverage Hamming distance and Lin1 in the methods based on dimensionality reduction in the performance comparison.

For those methods based on dimensionality reduction, we execute MDS/ tSNE/ UMAP 101 times to account for their stochasticity. We then use the median $p$-value (see Supplementary Table 2.1 and Supplementary Table 2.2) as the identification result of each data set (the detailed $p$-values of each method on ODSs and CRDSs are displayed in Supplementary Figure 4.1a and Supplementary Figure 4.2a). As demonstrated in Figure~\ref{fig:Count-Hamming}a and Supplementary Figure 2, several methods, including MDS Silv based on Hamming distance and Lin1, tSNE Silv based on Hamming distance, and MDS Dip based on Lin1, are capable of identifying all CRDSs correctly. However, these methods do not achieve a satisfactory level of success on identifying most ODSs as being clusterable, thereby revealing their potential limitations as clusterability evaluation methods. Moreover, as displayed in Supplementary Figure 4.1b and Supplementary Figure 4.2b, the methods based on dimensionality reduction generally tend to classify only six data sets, namely So, Zo, De, Hv, Ca, and Bc, as being clusterable.

For the transformation of categorical data into numerical data, it is important to note that such process might result in the loss of some information. As shown in our experiments (see Supplementary Table 3.1 and Supplementary Table 3.2), these embedding-based methods did not achieve better performance than the dimensionality reduction-based methods. As illustrated in Figure~\ref{fig:Count-Hamming}b and Supplementary Figure 3, they can identify the majority of ODSs as being clusterable, yet struggle to reliably recognize most CRDSs as being unclusterable. Furthermore, all embedding-based methods also fall short in identifying more than half of the target data sets as indicated by the intersection set.

\begin{figure}[!tbp]
	\centering
	\includegraphics[width=0.9\linewidth]{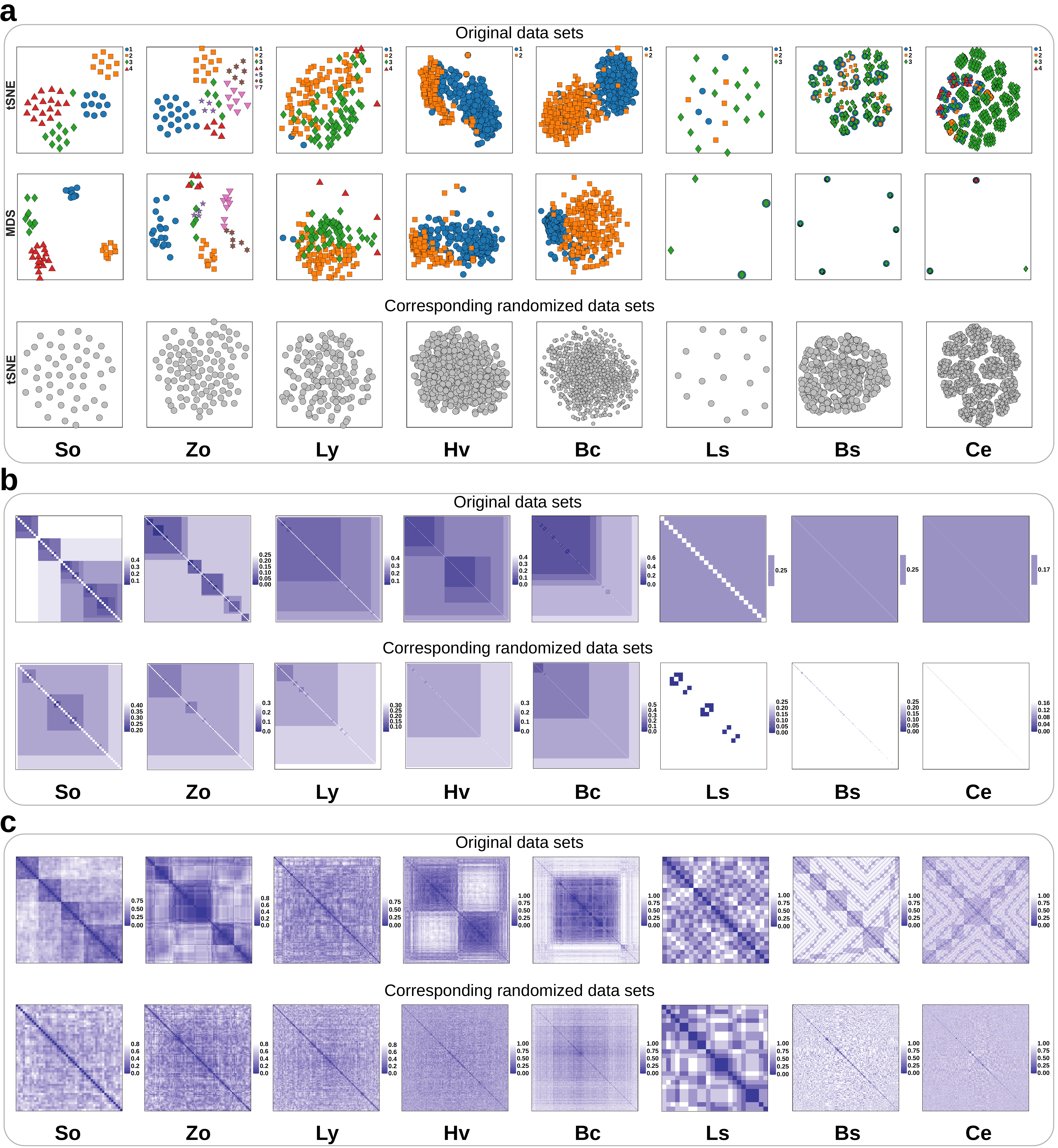} 
	\caption{Illustration of clustering structure underlying 8 UCI categorical data sets by using visual assessment. All plots are derived from the Hamming distances between objects in the data sets. Here, for each original data set, we generate its corresponding randomized data set, which undoubtedly should have no clustering structure (detailed procedures for generating the random data are presented in Section~\ref{methods:generation}). \textbf{(a)} Scatter plots of tSNE and MDS, where different colors/ shapes represent the class labels provided in the original data sets. Note that duplicate objects have been removed before running tSNE and MDS. Both \textbf{(b)} iVAT plots and \textbf{(c)} Dissimilarity plots display the reordered distances obtained through R Package ``seriation''. Potential clusters can be identified as multiple densely shaded blocks along the diagonal, where each square is large enough to accommodate a sufficient number of objects. The results from applying iVAT plots to the other 10 UCI categorical data sets are displayed in Supplementary Figure 5.} 
	\label{fig:VisualTools}
\end{figure}

In order to elucidate rationale of our validation strategy used above, we employ visualization tools to inspect different types of data sets. As depicted in Figure~\ref{fig:VisualTools}, we present the visualization results on eight data sets by using three tools: scatter plots in reduced two-dimensions space (Figure~\ref{fig:VisualTools}a), iVAT plots (Figure~\ref{fig:VisualTools}b), and Dissimilarity plots (Figure~\ref{fig:VisualTools}c). From the visual representations of ODSs and CRDSs, it can be observed that not all ODSs (e.g. Ls, Bs, Ce) has clearly observed clustering structure. This partially illustrate why our method regards the ODS of Ls, Bs and Ce as being unclusterable. The visualization results based on iVAT for the remaining ten data sets are provided in Supplementary Figure 5.

\subsection{Analysis of TestCat}
\label{results:analysis}

\subsubsection{Uniformity}
\begin{figure}[t]
	\centering
	\includegraphics[width=0.7\linewidth]{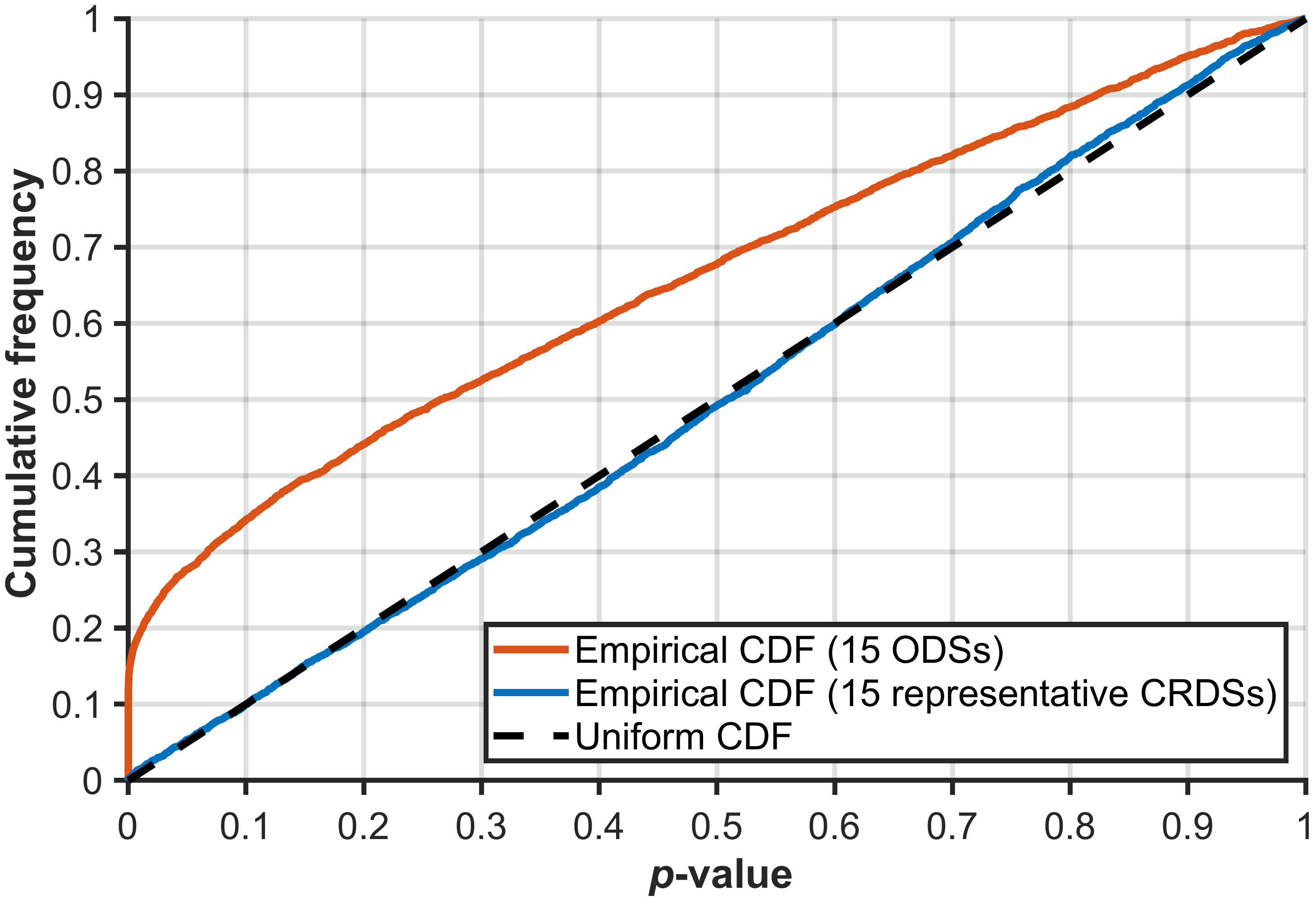} 
	\caption{Comparison of empirical CDFs for $p$-values of attribute pairs with a Uniform CDF. Such comparisons for each data set are provided in Supplementary Figure 6.} 
	\label{fig:uniformity}
\end{figure}

Our clusterability evaluation method, while differing from widely used methods for numerical data in terms of its null hypothesis and test statistic, still relies on the principle of Uniformity. For randomized categorical data, the association strength between any attribute pairs is expected to be uniformly distributed across the $[0, 1]$ interval, provided the association strength is arbitrary. This principle holds regardless of the data's scale, ensuring that the $p$-values for attribute pairs are unbiased and reflect random behavior when there is no conclusive evidence to suggest the presence of a cluster structure.
	
To validate this, we re-analyzed 15 of the 18 real-world data sets (excluding Ls, Bs, and Ce) that were identified as being clusterable by our TestCat in previous experiments. We compared the empirical cumulative distribution functions (CDFs) of these $p$-values (prior to aggregation into the final $p$-value) from the set of ODSs and their representative CRDSs to a theoretical Uniform CDF over the interval $[0, 1]$, as shown in~Figure \ref{fig:uniformity}. According to the Kolmogorov-Smirnov test, the CDF from the CRDSs showed no significant difference from the Uniform distribution at a significance level of 0.05 or 0.01. In contrast, the CDF from the ODSs exhibited significant deviations from Uniformity. This confirms that the $p$-values measuring the association strength between attribute pairs from randomized data are approximately uniformly distributed.
	
Furthermore, our TestCat accurately captures significant associations, aggregating from the attribute-pair level to the data set level, with the overall $p$-value potentially skewing toward smaller or larger values depending on the observed CDF. Specifically, when strong associations are present among attribute pairs, resulting in many CDF values gathering at the lower tail, as shown in~Figure \ref{fig:uniformity}, the TestCat derives an overall $p$-value that tends to skew toward smaller values. However, such skew is unlikely to occur when the associations among attribute pairs in randomized data are uniformly distributed, ensuring that extreme small final $p$-values are avoided.

\subsubsection{Robustness}

\begin{figure}[t]
	\centering
	\includegraphics[width=0.99\linewidth]{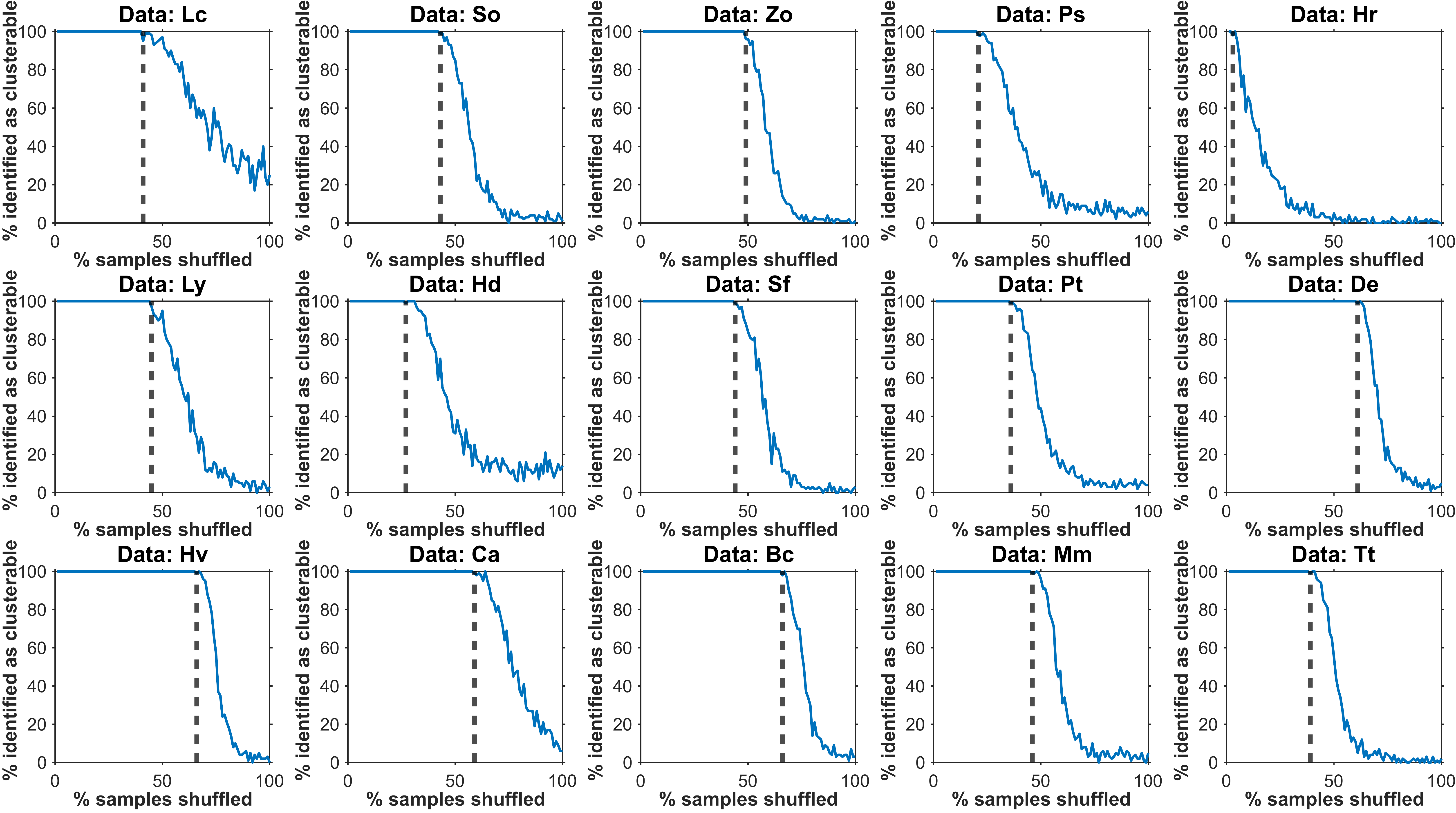} 
	\caption{Robustness of TestCat in determining clusterability against increasing randomness.} 
	\label{fig:against}
\end{figure}

A good clusterability evaluation method should demonstrate robustness across various randomized data sets, correctly identifying them as being unclusterable in most circumstances. Our approach exemplifies this property using the association strength among attributes. As illustrated in Supplementary Figure 1b, the majority of strongly associated attribute pairs fall within the relatively narrow range across the 101 CRDSs for each ODS. In contrast, the proportion of associated attribute pairs for clusterable ODS markedly exceed this value, displaying a broader distribution range. As a result, the $p$-value for randomized data sets consistently exceed that of the clusterable ODS, implying a clear distinction between the ODS and these randomized data sets (refer to Figure~\ref{fig:Metrics-cate}a).

As depicted in Supplementary Figure 7a, using a representative CRDS for each ODS, we observe a clear distinction in the number of positively associated attribute value pairs between the ODS and CRDS across eight data sets. These situations can be classified into two types: one where the number of associated pairs of the ODS markedly exceeds that of the CRDS, and another where the number of associated pairs of the ODS is zero (data sets Ls, Bs and Ce). The former scenario aligns with our criteria for identifying clusterable ODS, while the latter leads our method to compute a $p$-value of 1, marking them as unclusterable (an outcome that is consistent with the evidence provided by iVAT). Such plots on the number of associated pairs for the remaining data sets are displayed in Supplementary Figure 8. In all these cases, the number of associated pairs for the ODS is evidently greater than that for the CRDS, even though the difference is less pronounced in the data sets Lc and Ps. More specifically, as shown in Supplementary Figure 7b, the proportion of associated attribute pairs for clusterable ODSs exceeds 20\% in most cases. In contrast, the proportion for CRDSs is below 8.7\% in all cases. 

In addition to demonstrating TestCat's robustness in distinguishing between the clusterability of ODS and CRDS, we further extend our analysis to a simulation study examining TestCat's performance under conditions of partial randomness. We generate locally randomized data in the same manner as described in Section~\ref{methods:generation}, but instead of shuffling all objects across attributes, we shuffle only a fraction of them. As shown on the horizontal axis of Figure~\ref{fig:against}, we progressively shuffle 1\%, 2\%, and up to 100\% of the objects, with 100\% corresponding to the generation of a CRDS. Each shuffle is independently repeated 100 times, and the vertical axis represents the proportion of cases that TestCat identifies as being clusterable. The results indicate that when only a small fraction of objects are randomized, the derived $p$-values remain below the significance level of 0.01, identifying the cases as being clusterable with high accuracy. As more objects are shuffled, the strength of associations diminishes, and TestCat gradually fails to accurately detect clustering structures, classifying more cases as being unclusterable. When 100\% of the objects are shuffled (equivalent to generating CRDS), these cases are consistently identified as being unclusterable. Notably, even when nearly 50\% or more of the objects are shuffled, TestCat can still perfectly identify the cases as being clusterable with 100\% accuracy (as shown by the horizontal line in each plot of Figure~\ref{fig:against}) based on most of the clusterable data sets: Ls, So, Zo, Ly, Sf, De, Hv, Ca, Bc, Mm and Tt. This highlights TestCat's robustness against increasing randomness across a wide range of different clusterable data sets.

\begin{figure}[t]
	\centering
	\includegraphics[width=0.85\linewidth]{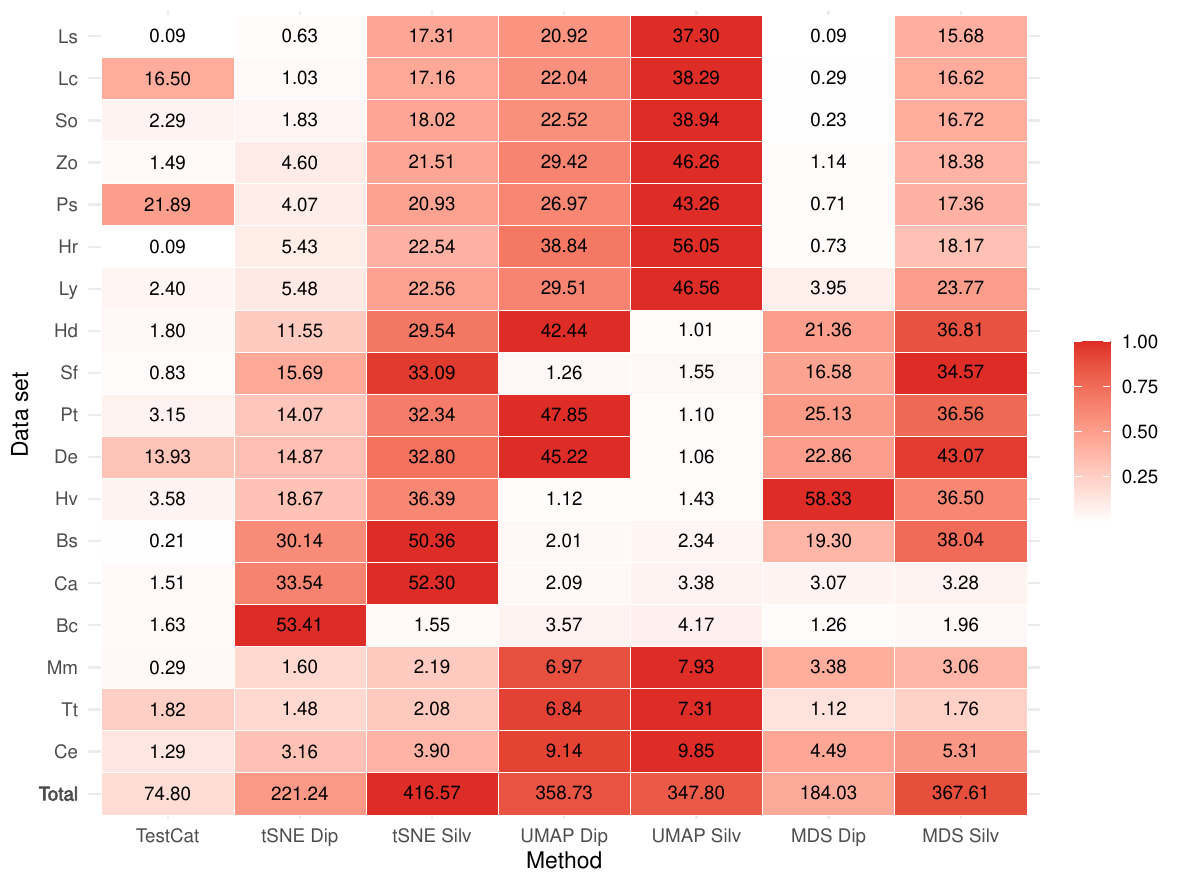} 
	\caption{Runtime for 50 iterations of each method across 18 UCI data sets. The values within each cell of the heatmap represent the actual runtime in seconds. The coloring of the heatmap is determined by the normalized runtime: for each data set, the runtime is scaled to the range (0,1] by dividing it by the maximum runtime. The corresponding color scale is illustrated in the legend on the right side.} 
	\label{fig:RT}
\end{figure}

\subsubsection{Efficiency}
Efficiency is crucial when a clusterability evaluation method is applied to large-scale data sets. To underscore this point, we assessed TestCat's runtime on 18 UCI data sets and compared it with other methods, specifically those based on dimensionality reduction. As illustrated in Figure~\ref{fig:RT}, TestCat showcased superior speed, topping the charts on eight specific data sets: Ls, Hr, Ly, Sf, Bs, Ca, Mm, and Ce. Notably, on 15 data sets, excluding Lc, Ps, and De, TestCat executed in less than 25\% of the duration required by the slowest competitor. Overall, in comparison to the fastest benchmark, MDS Dip, TestCat achieved a runtime that was approximately 40\% of its execution time.

\subsection{Code and data availability}
TestCat is available on Github (\url{https://github.com/hulianyu/TestCat}). Detailed guidelines for its implementation, as well as for the execution of the compared methods, are provided. Information on the parameters used in the codes of the compared methods is presented in Supplementary Note 1.

The complete collection of 18 UCI data sets employed in our analysis is publicly accessible at \url{https://archive.ics.uci.edu/datasets} (Attribute Type = Categorical), with their properties illustrated in Table~\ref{tab:UCI}. All utilized CRDSs, in addition to the numerical representations for each ODS and CRDS, can be found in the same Github repository hosting TestCat.

\section{Conclusions and Discussions}
\label{conclusions}
Clusterability evaluation plays a crucial role in cluster analysis. The importance of this step is underscored by its ability to authenticate the validity of clustering results, ensuring that identified patterns truly originate from the data rather than being mere by-product of the clustering process. More explicitly, if data are determined as being unclusterable, the use of any clustering algorithm would invariably lead to meaningless results.

To address the clusterability evaluation issue for categorical data, we introduce a testing-based method by employing the association relationship among categorical attributes. To the best of our knowledge, this is the first piece of work to tackle the clusterability evaluation problem for categorical data. Compared to existing solutions that are not specially developed for categorical data, our TestCat approach not only demonstrates superior performance but also solves such a challenging issue in an intuitive, simple and elegant manner.

It is undeniable that our method harbors certain limitations: (1) The independence assumption~\cite{Ferrari2019} among attribute pairs, as well as the use of the chi-squared approximation (further discussed in Supplementary Note 2), generally do not hold. If we do not adhere to these assumptions, the deviation of an analytical $p$-value for significance assessment would be a challenging issue. (2) The chi-squared test may not be the best choice for association testing in the context of clusterability evaluation, we need to further check other more appropriate methods. (3) Association testing may lack validity for data structures with intricate attribute interactions. For instance, in the Bs (Balance Scale) data set, cluster formation may rely on multi-attribute interactions beyond just pairwise associations. Additionally, in the Ce (Car Evaluation) data set, the absence of associations between individual attributes does not necessarily imply there are no clusters, as these hierarchical attributes could be strongly associated with a cluster variable. These scenarios, as detailed in Supplementary Note 3, indicate the need for developing alternative methods capable of capturing such complex relationships.

Finally, our research efforts in this article hint at several aspects. First, it highlights the importance of the clusterability evaluation issue for categorical data, which is overlooked during the past decades. Second, we empirically demonstrate the fact that existing solutions for numeric data cannot solve this problem very well, indicating that it is still necessary to develop new algorithms towards this direction.  Finally, there are many directions for future research, ranging from the refinement of TestCat to the development of brand-new methods.

\bmhead{Acknowledgments}
This work has been partially supported by the Natural Science Foundation of China under Grant No. 62472064.


\bibliography{sn-bibliography}

\end{document}


\author{Lianyu Hu}
\affiliation{
	School of Software, Dalian University of Technology, Dalian, 116024, China}

\author{Junjie Dong}
\affiliation{
	School of Software, Dalian University of Technology, Dalian, 116024, China}
\author{Mudi Jiang}
\affiliation{
	School of Software, Dalian University of Technology, Dalian, 116024, China}
\author{Yan Liu}
\affiliation{
	School of Software Engineering, Dalian University, Dalian, 116622, China}

\author{Zengyou He}
\email{zyhe@dlut.edu.cn}
\affiliation{
	School of Software, Dalian University of Technology, Dalian, 116024, China}
\affiliation{
	Key Laboratory for Ubiquitous Network and Service Software of Liaoning Province, Dalian, 116024, China}

\title{Supplementary Information for \\
	``Clusterability test for categorical data''}

\maketitle


\section*{Supplementary Method}
\textbf{1. Standardized residuals:}
As a by-product of chi-squared tests, standardized residuals ($sr$) offer a granular measure of association strength between two attribute values. For the $s$-th attribute pair, specifically between attribute values $A_i$ and $B_j$, the standardized residual, $sr^{(s)}_{ij}$, is calculated as follows:
\begin{equation}
	sr^{(s)}_{ij} = \frac{O^{(s)}_{ij}-E^{(s)}_{ij}}{\sqrt{E^{(s)}_{ij}(1-\frac{O^{(s)}_{i\cdot}}{O^{(s)}_{\cdot \cdot}})(1-\frac{O^{(s)}_{\cdot j}}{O^{(s)}_{\cdot \cdot}})}} .
\end{equation}

\section*{Supplementary Note}
\textbf{1. The parameters used in the codes of compared methods:} \\

The Dip and Silverman Tests functions can be accessed from the \texttt{dipsilverman.R} file found in the package source of ``clusterability'' on CRAN: \url{https://cran.r-project.org/web/packages/clusterability}. When applying the Dip and Silverman Tests, the following parameters are utilized:
\begin{verbatim}
	# `fit' denotes the (transformed) one-dimensional values
	source(`dipsilverman.R')
	dip(fit, simulatepvalue = FALSE, B = 2000) 
	silverman(fit, k=1, M=999, adjust=TRUE)
\end{verbatim}

In addition, we also make use of various dimensionality reduction techniques including MDS, tSNE, UMAP, as well as PCA and SPCA. The specific parameters for these methods are as follows:
\begin{verbatim}
	# `dist' is a distance matrix
	# `y' is a (embedding) numerical data set
	
	# MDS
	library(smacof) 
	fit <- mds(dist, ndim = 1, type="ordinal", init = "random", ties="secondary")$conf
	fit <- scale(fit) # Scale the data
	# tSNE
	library(Rtsne) 
	pt <- min(floor((nrow(dist) - 1) / 3), 30)
	fit <- Rtsne(dist, dims=1, perplexity = pt, is_distance = TRUE)$Y
	fit <- scale(fit) # Scale the data
	# UMAP
	library(uwot)
	fit <- umap(dist, n_components=1)
	fit <- scale(fit) # Scale the data
	# PCA
	fit <- stats::prcomp(y, center = TRUE, scale. = FALSE, retx = TRUE)$x[, 1] 
	# SPCA
	library(sparsepca)
	fit <- spca(y, k = 1)$scores
\end{verbatim}

\textbf{2. Monte Carlo simulation for $p$-value calculation:} \\

To accurately derive a $p$-value from an asymptotic chi-squared distribution, each cell in the contingency table (CT) needs a sufficiently large expected frequency. However, not all the data sets we used meet this criterion. Additionally, since our approach involves the summation of chi-squared values across all attributes, it does not correspond to a specific contingency table for each cell. This differs from the scenario when dealing with a single pair of attributes.

To achieve a more precise $p$-value calculation in such scenarios, we adopt sampling methods like the Monte Carlo test, as recommended by Hope (1968). Specifically, we utilize the R function \texttt{chisq.test(CT, simulate.p.value = TRUE, B = 20000)} to compute the $p$-value for each pair of attributes. This is done using the Monte Carlo method with a large number of replicates, set at 20,000 in our case. We then aggregate all these individual $p$-values into a single composite value using Fisher's method (implemented in the R Package ``poolr''), resulting in what we consider a more refined estimation of the $p$-value.

For reference, the $p$-values derived via the Monte Carlo method are provided in Supplementary Table 4. The results show consistency with our original conclusions regarding significance for both the ODS and CRDS data sets. However, it is important to note that deriving $p$-values through Monte Carlo simulations can be time-consuming due to the large volume of simulated data required. Therefore, unless there is a significant improvement in the results, we will continue using our original computational method.\\

\textbf{3. Special case analysis of Bs and Ce data sets} \\

(1) Emergence of clusters from multi-attribute interactions in the \textbf{Bs} data set: The Bs data set, comprising four ordinal attributes (Left-Weight, Left-Distance, Right-Weight, and Right-Distance), illustrates how clusters in high-dimensional data can emerge from interactions involving more than just two attributes. Class labels (L, B, R) are derived from the linear formula $f = \text{Left-Weight}\times \text{Left-Distance} - \text{Right-Weight} \times \text{Right-Distance}$. Objects are classified as R, B, or L depending on whether $f$ is less than, equal to, or greater than zero, respectively.\\

(2) Emergence of clusters from hierarchical attributes in the \textbf{Ce} data set: The Ce presents a different challenge. Our method deemed it to be unclusterable based on paired inter-attribute associations. However, class labels in this data set may arise from hierarchical attribute structures, not effectively captured by our current approach. Consider the following example:
\begin{center}
	\begin{forest}
		for tree={
			draw,
			align=center,
			parent anchor=south,
			child anchor=north,
			edge={->,>=latex},
			l sep+=0.2cm,
			s sep+=0.2cm,
		}
		[Car Evaluation
		[Ownership Costs
		[buying \\
		(vhigh/high/med/low)
		]
		[maint\\
		(vhigh/high/med/low)
		]
		]
		[Experience
		[Comfort
		[doors\\
		(2/3/4/5more)
		]
		[persons\\
		(2/4/more)
		]
		[lug\_boot \\
		(small/med/big)
		]
		]
		[safety\\
		(low/med/high)
		]
		]
		]
	\end{forest}
\end{center}
This example illustrates the existence of attributes at different hierarchical levels, and it is inappropriate to compare attributes across these varied levels. Even within the same hierarchical level, attributes could represent completely distinct objective factors. For example, high Ownership Costs do not necessarily associate with the Experience being good or bad. These factors independently influence the class labels, leading to strong associations between certain attributes and class labels. Nonetheless, when all inter-attribute comparisons are pooled together, these associations may seem unrelated.

\newpage
\section*{Supplementary Tables}
\newcolumntype{Y}{>{\centering\arraybackslash}X}

\begin{table}[h]
	\caption*{\textbf{Supplementary Table 1}: Percentage (\%) of occurrences where the $p$-value exceeds 0.01 across \textbf{the 101 runs} (calculated by TestCat). The distribution of these $p$-values for each data set can be found in Supplementary Figure 1a.} 
	\centering
	\begin{tabularx}{\textwidth}{|c|YYYYYYYYYYYYYYYYYY|}
		\hline
		CRDS & Ls & Lc & So & Zo & Ps & Hr & Ly & Hd & Sf & Pt & De & Hv & Bs & Ca & Bc & Mm & Tt & Ce\\
		\hline
		$p\text{-value}>0.01$ & 99 & 78 & 98 & 99 & 97 & 98 & 96 & 88 & 100 & 93 & 97 & 100 & 100 & 92 & 97 & 99 & 99 & 100\\
		\hline
	\end{tabularx}
\end{table}

\begin{table}[h]
	\caption*{\textbf{Supplementary Table 2.1}: The clusterability analysis results of Dip and Silverman via dimensionality reduction based on \textbf{Hamming} distance.  All results are represented by the median $p$-value obtained from 101 runs for each ODS or CRDS.} 
	\centering
		\begin{tabularx}{\textwidth}{|Y|YY|YY|YY|YY|YY|YY|}
			\hline
			& \multicolumn{2}{c|}{\textbf{MDS Dip}} & \multicolumn{2}{c|}{\textbf{MDS Silv}}  & \multicolumn{2}{c|}{\textbf{tSNE Dip}} & \multicolumn{2}{c|}{\textbf{tSNE Silv}} & \multicolumn{2}{c|}{\textbf{UMAP Dip}} &  \multicolumn{2}{c|}{\textbf{UMAP Sil}}  \\
			\cline{2-13}
			\multicolumn{1}{|c|}{\multirow{-2}{*}{\centering Data set}}  & \textcolor{dr}{\textbf{ODS}} & \textcolor{db}{\textbf{CRDS}} & \textcolor{dr}{\textbf{ODS}} & \textcolor{db}{\textbf{CRDS}} & \textcolor{dr}{\textbf{ODS}} & \textcolor{db}{\textbf{CRDS}} & \textcolor{dr}{\textbf{ODS}} & \textcolor{db}{\textbf{CRDS}} & \textcolor{dr}{\textbf{ODS}} & \textcolor{db}{\textbf{CRDS}} & \textcolor{dr}{\textbf{ODS}} & \textcolor{db}{\textbf{CRDS}} \\
			\textbf{Ls} & 0.99961          & \cellcolor{db}\textcolor{white}{\textbf{0.99403}} & 0.59925          & \cellcolor{db}\textcolor{white}{\textbf{0.50165}} & 1                & \cellcolor{db}\textcolor{white}{\textbf{1}}       & 0.98736          & \cellcolor{db}\textcolor{white}{\textbf{0.99195}} & 0.98365          & \cellcolor{db}\textcolor{white}{\textbf{0.11939}} & 0.24719          & 0.00038          \\
			\textbf{Lc} & 0.98938          & \cellcolor{db}\textcolor{white}{\textbf{0.99697}} & 0.34771          & \cellcolor{db}\textcolor{white}{\textbf{0.56481}} & 1                & \cellcolor{db}\textcolor{white}{\textbf{1}}       & 0.99080          & \cellcolor{db}\textcolor{white}{\textbf{0.99080}} & 0.99280          & \cellcolor{db}\textcolor{white}{\textbf{0.99468}} & 0.57284          & \cellcolor{db}\textcolor{white}{\textbf{0.83579}} \\
			\textbf{So} & 0.99751          & \cellcolor{db}\textcolor{white}{\textbf{0.99985}} & 0.32464          & \cellcolor{db}\textcolor{white}{\textbf{0.54529}} & \cellcolor{lr}\textbf{0.00469} & \cellcolor{db}\textcolor{white}{\textbf{0.90943}} & 0.09412          & \cellcolor{db}\textcolor{white}{\textbf{0.61648}} & \cellcolor{lr}\textbf{0.00040} & \cellcolor{db}\textcolor{white}{\textbf{0.99437}} & \cellcolor{lr}\textbf{0.00101} & \cellcolor{db}\textcolor{white}{\textbf{0.58662}} \\
			\textbf{Zo} & 1                & \cellcolor{db}\textcolor{white}{\textbf{0.99459}} & 0.39142          & \cellcolor{db}\textcolor{white}{\textbf{0.83809}} & \cellcolor{lr}\textbf{0.00043} & \cellcolor{db}\textcolor{white}{\textbf{0.53732}} & \cellcolor{lr}\textbf{0.00385} & \cellcolor{lb}\textcolor{white}{\textbf{0.05239}} & \cellcolor{dr}\textbf{0}       & \cellcolor{db}\textcolor{white}{\textbf{0.98524}} & \cellcolor{dr}\textbf{0}       & \cellcolor{db}\textcolor{white}{\textbf{0.11559}} \\
			\textbf{Ps} & 1  & \cellcolor{db}\textcolor{white}{\textbf{1}} & 0.57399 & \cellcolor{db}\textcolor{white}{\textbf{0.50280}}\ & 0.66863 & \cellcolor{db}\textcolor{white}{\textbf{0.72726}} & 0.58433 & \cellcolor{db}\textcolor{white}{\textbf{0.42128}} & 0.99296 & \cellcolor{db}\textcolor{white}{\textbf{0.99675}} & 0.58662 & \cellcolor{db}\textcolor{white}{\textbf{0.88861}} \\
			\textbf{Hr} & 1 & \cellcolor{db}\textcolor{white}{\textbf{1}}      & 0.55447          & \cellcolor{db}\textcolor{white}{\textbf{0.59236}} & 0.08859          & \cellcolor{db}\textcolor{white}{\textbf{0.15269}} & 0.09064          & \cellcolor{db}\textcolor{white}{\textbf{0.28770}} & 0.05615          & \cellcolor{db}\textcolor{white}{\textbf{0.79364}} & 0.01457          & \cellcolor{lb}\textcolor{white}{\textbf{0.03410}} \\
			\textbf{Ly} & 0.99598          & \cellcolor{db}\textcolor{white}{\textbf{0.99838}} & 0.54299          & \cellcolor{db}\textcolor{white}{\textbf{0.64404}} & 0.70795          & \cellcolor{db}\textcolor{white}{\textbf{0.74649}} & 0.33503          & \cellcolor{db}\textcolor{white}{\textbf{0.29229}} & 0.79781          & \cellcolor{db}\textcolor{white}{\textbf{0.99481}} & 0.09480          & \cellcolor{db}\textcolor{white}{\textbf{0.75771}} \\
			\textbf{Hd} & 0.99708          & \cellcolor{db}\textcolor{white}{\textbf{1}}      & 0.22919          & \cellcolor{db}\textcolor{white}{\textbf{0.95521}} & 0.18627          & \cellcolor{db}\textcolor{white}{\textbf{0.66120}} & \cellcolor{lr}\textbf{0.00518} & \cellcolor{db}\textcolor{white}{\textbf{0.26633}} & 0.09117          & \cellcolor{db}\textcolor{white}{\textbf{0.99362}} & \cellcolor{lr}\textbf{0.00038} & \cellcolor{db}\textcolor{white}{\textbf{0.62337}} \\
			\textbf{Sf} & \cellcolor{lr}\textbf{0.00143}          & \cellcolor{db}\textcolor{white}{\textbf{0.54325}} & 0.19291          & \cellcolor{db}\textcolor{white}{\textbf{0.78297}} & 0.01240          & \cellcolor{db}\textcolor{white}{\textbf{0.18430}} & 0.14020          & \cellcolor{db}\textcolor{white}{\textbf{0.11925}} & \cellcolor{lr}\textbf{0.00033}          & \cellcolor{db}\textcolor{white}{\textbf{0.12846}} & \cellcolor{lr}\textbf{0.00002} & \cellcolor{lb}\textcolor{white}{\textbf{0.01141}} \\
			\textbf{Pt} & 0.99353          & \cellcolor{db}\textcolor{white}{\textbf{1}}       & 0.98047          & \cellcolor{db}\textcolor{white}{\textbf{0.85876}} & 0.42551          & \cellcolor{db}\textcolor{white}{\textbf{0.54549}} & 0.19921          & \cellcolor{db}\textcolor{white}{\textbf{0.49132}} & \cellcolor{lr}\textbf{0.00161}         & \cellcolor{db}\textcolor{white}{\textbf{0.98143}} & \cellcolor{lr}\textbf{0.00350} & \cellcolor{db}\textcolor{white}{\textbf{0.25766}} \\
			\textbf{De} & 0.68675          & \cellcolor{db}\textcolor{white}{\textbf{1}}      & 0.02406          & \cellcolor{db}\textcolor{white}{\textbf{0.99654}} & \cellcolor{dr}\textbf{0}       & \cellcolor{db}\textcolor{white}{\textbf{0.73729}} & \cellcolor{dr}\textbf{0}       & \cellcolor{db}\textcolor{white}{\textbf{0.48443}} & \cellcolor{dr}\textbf{0}       & \cellcolor{db}\textcolor{white}{\textbf{0.99712}} & \cellcolor{dr}\textbf{0}       & \cellcolor{db}\textcolor{white}{\textbf{0.90813}} \\
			\textbf{Hv} & \cellcolor{lr}\textbf{0.00299} & \cellcolor{db}\textcolor{white}{\textbf{1}}      & \cellcolor{lr}\textbf{0.00248} & \cellcolor{db}\textcolor{white}{\textbf{0.46721}} & \cellcolor{lr}\textbf{0.00001} & \cellcolor{db}\textcolor{white}{\textbf{0.84918}} & \cellcolor{dr}\textbf{0}       & \cellcolor{db}\textcolor{white}{\textbf{0.58892}} & \cellcolor{dr}\textbf{0}       & \cellcolor{db}\textcolor{white}{\textbf{0.99549}} & \cellcolor{dr}\textbf{0}       & \cellcolor{db}\textcolor{white}{\textbf{0.77953}} \\
			\textbf{Bs} & 1  & \cellcolor{db}\textcolor{white}{\textbf{1}}    & 0.77493          & \cellcolor{db}\textcolor{white}{\textbf{0.66585}} & \cellcolor{dr}\textbf{0}       & \cellcolor{db}\textcolor{white}{\textbf{0.50439}} & 0.04431          & \cellcolor{db}\textcolor{white}{\textbf{0.52003}} & 0.08893          & \cellcolor{db}\textcolor{white}{\textbf{0.84963}} & 0.35462          & \cellcolor{db}\textcolor{white}{\textbf{0.43965}} \\
			\textbf{Ca} & 0.01624          & \cellcolor{db}\textcolor{white}{\textbf{0.45886}} & \cellcolor{lr}\textbf{0.00461} & \cellcolor{db}\textcolor{white}{\textbf{0.47295}} & \cellcolor{lr}\textbf{0.00289} & \cellcolor{db}\textcolor{white}{\textbf{0.25492}} & \cellcolor{lr}\textbf{0.00385} & \cellcolor{lb}\textcolor{white}{\textbf{0.05386}} & \cellcolor{dr}\textbf{0}       & 0.00042          & \cellcolor{dr}\textbf{0}       & 0                \\
			\textbf{Bc} & 0.03966          & \cellcolor{db}\textcolor{white}{\textbf{1}}       & \cellcolor{lr}\textbf{0.00001} & \cellcolor{db}\textcolor{white}{\textbf{0.97702}} & 0.10325          & \cellcolor{db}\textcolor{white}{\textbf{0.36668}} & \cellcolor{lr}\textbf{0.00932} & \cellcolor{db}\textcolor{white}{\textbf{0.36268}} & \cellcolor{dr}\textbf{0}       & \cellcolor{db}\textcolor{white}{\textbf{0.99407}} & \cellcolor{dr}\textbf{0}       & \cellcolor{db}\textcolor{white}{\textbf{0.60155}} \\
			\textbf{Mm} & \cellcolor{dr}\textbf{0}       & 0.00093          & \cellcolor{dr}\textbf{0}       & \cellcolor{lb}\textcolor{white}{\textbf{0.03135}} & \cellcolor{dr}\textbf{0}       & 0                & \cellcolor{lr}\textbf{0.00001} & \cellcolor{db}\textcolor{white}{\textbf{0.39946}} & \cellcolor{dr}\textbf{0}       & 0                & \cellcolor{dr}\textbf{0}       & 0                \\
			\textbf{Tt} & 1                & \cellcolor{db}\textcolor{white}{\textbf{1}}     & 0.65667          & \cellcolor{db}\textcolor{white}{\textbf{0.54644}} & 0.17007          & \cellcolor{db}\textcolor{white}{\textbf{0.46618}} & 0.60155          & \cellcolor{db}\textcolor{white}{\textbf{0.34079}} & 0.72710          & \cellcolor{db}\textcolor{white}{\textbf{0.99451}} & 0.70948          & \cellcolor{db}\textcolor{white}{\textbf{0.59351}} \\
			\textbf{Ce} & 1                & \cellcolor{db}\textcolor{white}{\textbf{1}}     & 0.81857          & \cellcolor{db}\textcolor{white}{\textbf{0.76919}} & \cellcolor{dr}\textbf{0}       & \cellcolor{lb}\textcolor{white}{\textbf{0.08486}} & 0.02626          & \cellcolor{lb}\textcolor{white}{\textbf{0.02363}} & \cellcolor{lr}\textbf{0.00004} & \cellcolor{lb}\textcolor{white}{\textbf{0.01389}} & \cellcolor{lr}\textbf{0.00806} & 0.00849     \\
			\hline
		\end{tabularx}
\end{table}

\begin{table}[h]
	\caption*{\textbf{Supplementary Table 2.2}: The clusterability analysis results of Dip and Silverman via dimensionality reduction based on \textbf{Lin1} distance.  All results are represented by the median $p$-value obtained from 101 runs for each ODS or CRDS.} 
	\centering
	\begin{tabularx}{\textwidth}{|Y|YY|YY|YY|YY|YY|YY|}
		\hline
		& \multicolumn{2}{c|}{\textbf{MDS Dip}} & \multicolumn{2}{c|}{\textbf{MDS Silv}}  & \multicolumn{2}{c|}{\textbf{tSNE Dip}} & \multicolumn{2}{c|}{\textbf{tSNE Silv}} & \multicolumn{2}{c|}{\textbf{UMAP Dip}} &  \multicolumn{2}{c|}{\textbf{UMAP Sil}}  \\
		\cline{2-13}
		\multicolumn{1}{|c|}{\multirow{-2}{*}{\centering Data set}} & \textcolor{dr}{\textbf{ODS}} & \textcolor{db}{\textbf{CRDS}} & \textcolor{dr}{\textbf{ODS}} & \textcolor{db}{\textbf{CRDS}} & \textcolor{dr}{\textbf{ODS}} & \textcolor{db}{\textbf{CRDS}} & \textcolor{dr}{\textbf{ODS}} & \textcolor{db}{\textbf{CRDS}} & \textcolor{dr}{\textbf{ODS}} & \textcolor{db}{\textbf{CRDS}} & \textcolor{dr}{\textbf{ODS}} & \textcolor{db}{\textbf{CRDS}} \\
		\textbf{Ls} & 1                & \cellcolor{db}\textcolor{white}{\textbf{1}}       & 1          & \cellcolor{db}\textcolor{white}{\textbf{1}}       & 1                & \cellcolor{db}\textcolor{white}{\textbf{1}}       & 1                & \cellcolor{db}\textcolor{white}{\textbf{1}}       & 1                & \cellcolor{db}\textcolor{white}{\textbf{1}}       & 1                & \cellcolor{db}\textcolor{white}{\textbf{1}}       \\
		\textbf{Lc} & 0.99392          & \cellcolor{db}\textcolor{white}{\textbf{0.99762}} & 0.47984    & \cellcolor{db}\textcolor{white}{\textbf{0.46721}} & 1                & \cellcolor{db}\textcolor{white}{\textbf{1}}       & 0.99310          & \cellcolor{db}\textcolor{white}{\textbf{0.99310}} & 0.99115          & \cellcolor{db}\textcolor{white}{\textbf{0.99398}} & 0.27747          & \cellcolor{db}\textcolor{white}{\textbf{0.73360}} \\
		\textbf{So} & 0.99965          & \cellcolor{db}\textcolor{white}{\textbf{1}}      & 0.49362    & \cellcolor{db}\textcolor{white}{\textbf{0.47410}} & \cellcolor{lr}\textbf{0.00970} & \cellcolor{db}\textcolor{white}{\textbf{0.85177}} & 0.01141          & \cellcolor{db}\textcolor{white}{\textbf{0.69341}} & \cellcolor{dr}\textbf{0}       & \cellcolor{db}\textcolor{white}{\textbf{0.99449}} & \cellcolor{lr}\textbf{0.00001} & \cellcolor{db}\textcolor{white}{\textbf{0.67274}} \\
		\textbf{Zo} & 1                & \cellcolor{db}\textcolor{white}{\textbf{1}}       & 0.43620    & \cellcolor{db}\textcolor{white}{\textbf{0.56021}} & \cellcolor{lr}\textbf{0.00013} & \cellcolor{db}\textcolor{white}{\textbf{0.58896}} & \cellcolor{lr}\textbf{0.00002} & \cellcolor{lb}\textcolor{white}{\textbf{0.09412}} & \cellcolor{dr}\textbf{0}       & \cellcolor{db}\textcolor{white}{\textbf{0.99285}} & \cellcolor{lr}\textbf{0.00001} & \cellcolor{db}\textcolor{white}{\textbf{0.40176}} \\
		\textbf{Ps} & 1                & \cellcolor{db}\textcolor{white}{\textbf{1}}       & 0.50854    & \cellcolor{db}\textcolor{white}{\textbf{0.47524}} & 0.69424          & \cellcolor{db}\textcolor{white}{\textbf{0.87702}} & 0.41668          & \cellcolor{db}\textcolor{white}{\textbf{0.57514}} & 0.99312          & \cellcolor{db}\textcolor{white}{\textbf{0.99685}} & 0.59236          & \cellcolor{db}\textcolor{white}{\textbf{0.87483}} \\
		\textbf{Hr} & 0.94345          & \cellcolor{db}\textcolor{white}{\textbf{0.91676}} & 0.42932    & \cellcolor{db}\textcolor{white}{\textbf{0.46836}} & \cellcolor{dr}\textbf{0}       & 0                & \cellcolor{dr}\textbf{0}       & 0.00002          & \cellcolor{lr}\textbf{0.00001} & 0                & \cellcolor{dr}\textbf{0}       & 0.00038          \\
		\textbf{Ly} & 0.99453          & \cellcolor{db}\textcolor{white}{\textbf{1}}       & 0.24822    & \cellcolor{db}\textcolor{white}{\textbf{0.53840}} & 0.34873          & \cellcolor{db}\textcolor{white}{\textbf{0.86084}} & 0.03950          & \cellcolor{db}\textcolor{white}{\textbf{0.58662}} & 0.21797          & \cellcolor{db}\textcolor{white}{\textbf{0.99406}} & \cellcolor{lr}\textbf{0.00001} & \cellcolor{db}\textcolor{white}{\textbf{0.61074}} \\
		\textbf{Hd} & 1                & \cellcolor{db}\textcolor{white}{\textbf{1}}       & 0.36728    & \cellcolor{db}\textcolor{white}{\textbf{0.53955}} & 0.18538          & \cellcolor{db}\textcolor{white}{\textbf{0.84674}} & \cellcolor{lr}\textbf{0.00002} & \cellcolor{db}\textcolor{white}{\textbf{0.66356}} & 0.37514          & \cellcolor{db}\textcolor{white}{\textbf{0.99462}} & \cellcolor{dr}\textbf{0}       & \cellcolor{db}\textcolor{white}{\textbf{0.75312}} \\
		\textbf{Sf} & 0.48496          & \cellcolor{db}\textcolor{white}{\textbf{0.88428}}\ & 0.12575    & \cellcolor{db}\textcolor{white}{\textbf{0.42013}} & \cellcolor{lr}\textbf{0.00838} & \cellcolor{db}\textcolor{white}{\textbf{0.56854}} & 0.06150          & \cellcolor{db}\textcolor{white}{\textbf{0.64289}} & \cellcolor{lr}\textbf{0.00037} & \cellcolor{lb}\textcolor{white}{\textbf{0.04883}} & \cellcolor{dr}\textbf{0}       & 0.00001          \\
		\textbf{Pt} & 0.99156          & \cellcolor{db}\textcolor{white}{\textbf{0.99778}} & 0.32695    & \cellcolor{db}\textcolor{white}{\textbf{0.66126}} & 0.51133          & \cellcolor{db}\textcolor{white}{\textbf{0.48334}} & 0.23412          & \cellcolor{db}\textcolor{white}{\textbf{0.14218}} & 0.45254          & \cellcolor{db}\textcolor{white}{\textbf{0.96254}} & 0.02453          & \cellcolor{db}\textcolor{white}{\textbf{0.36037}} \\
		\textbf{De} & 0.77978          & \cellcolor{db}\textcolor{white}{\textbf{1}}       & 0.03648    & \cellcolor{db}\textcolor{white}{\textbf{0.40061}} & \cellcolor{dr}\textbf{0}       & \cellcolor{db}\textcolor{white}{\textbf{0.78075}} & \cellcolor{dr}\textbf{0}       & \cellcolor{db}\textcolor{white}{\textbf{0.41439}} & \cellcolor{dr}\textbf{0}       & \cellcolor{db}\textcolor{white}{\textbf{0.99675}} & \cellcolor{dr}\textbf{0}       & \cellcolor{db}\textcolor{white}{\textbf{0.85646}} \\
		\textbf{Hv} & 0.99950           & \cellcolor{db}\textcolor{white}{\textbf{1}}       & 0.24413    & \cellcolor{db}\textcolor{white}{\textbf{0.56251}} & \cellcolor{lr}\textbf{0.00024} & \cellcolor{db}\textcolor{white}{\textbf{0.38244}} & \cellcolor{dr}\textbf{0}       & \cellcolor{db}\textcolor{white}{\textbf{0.19502}} & \cellcolor{dr}\textbf{0}       & \cellcolor{db}\textcolor{white}{\textbf{0.99492}} & \cellcolor{dr}\textbf{0}       & \cellcolor{db}\textcolor{white}{\textbf{0.73360}} \\
		\textbf{Bs} & 1                & \cellcolor{db}\textcolor{white}{\textbf{1}}       & 1          & \cellcolor{db}\textcolor{white}{\textbf{1}}       & 1                & \cellcolor{db}\textcolor{white}{\textbf{1}}       & 1                & \cellcolor{db}\textcolor{white}{\textbf{1}}       & 1                & \cellcolor{db}\textcolor{white}{\textbf{1}}       & 1                & \cellcolor{db}\textcolor{white}{\textbf{1}}       \\
		\textbf{Ca} & 0.66305          &\cellcolor{db}\textcolor{white}{\textbf{0.99622}} & 0.02216    & \cellcolor{db}\textcolor{white}{\textbf{0.58662}} & 0.21438          & \cellcolor{db}\textcolor{white}{\textbf{0.64294}} & 0.14218          & \cellcolor{db}\textcolor{white}{\textbf{0.16608}} & \cellcolor{dr}\textbf{0}       & \cellcolor{db}\textcolor{white}{\textbf{0.28694}} & \cellcolor{dr}\textbf{0}       & \cellcolor{lb}\textcolor{white}{\textbf{0.02287}} \\
		\textbf{Bc} & \cellcolor{lr}\textbf{0.00654} & \cellcolor{db}\textcolor{white}{\textbf{1}}       & \cellcolor{dr}\textbf{0} & \cellcolor{db}\textcolor{white}{\textbf{0.99540}} & \cellcolor{dr}\textbf{0}       & \cellcolor{db}\textcolor{white}{\textbf{0.76192}} & \cellcolor{dr}\textbf{0}       & \cellcolor{db}\textcolor{white}{\textbf{0.40405}} & \cellcolor{dr}\textbf{0}       & \cellcolor{db}\textcolor{white}{\textbf{0.89109}} & \cellcolor{dr}\textbf{0}       & \cellcolor{db}\textcolor{white}{\textbf{0.20436}} \\
		\textbf{Mm} & \cellcolor{dr}\textbf{0}       & \cellcolor{lb}\textcolor{white}{\textbf{0.01101}} & \cellcolor{dr}\textbf{0} & \cellcolor{db}\textcolor{white}{\textbf{0.16075}} & \cellcolor{dr}\textbf{0}       & 0                & \cellcolor{dr}\textbf{0}       & 0.00057          & \cellcolor{dr}\textbf{0}       & 0                & \cellcolor{dr}\textbf{0}       & 0.00002          \\
		\textbf{Tt} & 1                & \cellcolor{db}\textcolor{white}{\textbf{1}}       & 0.61533    & \cellcolor{db}\textcolor{white}{\textbf{0.94947}} & 0.56673          & \cellcolor{db}\textcolor{white}{\textbf{0.49809}} & 0.01073          & \cellcolor{lb}\textcolor{white}{\textbf{0.02505}} & 0.78506          & 0.00695          & 0.07279          & 0                \\
		\textbf{Ce} & 1                & \cellcolor{db}\textcolor{white}{\textbf{1}}       & 1          & \cellcolor{db}\textcolor{white}{\textbf{1}}      & 1                & \cellcolor{db}\textcolor{white}{\textbf{1}}       & 1                & \cellcolor{db}\textcolor{white}{\textbf{1}}       & 1                & \cellcolor{db}\textcolor{white}{\textbf{1}}       & 1                & \cellcolor{db}\textcolor{white}{\textbf{1}}      \\
		\hline
	\end{tabularx}
\end{table}

\begin{table}[h]
	\caption*{\textbf{Supplementary Table 3.1}: The results of clusterability evaluation methods utilizing \textbf{CDC\_DR} (spectral embedding version) on 18 UCI data sets.} 
	\centering
		\begin{tabularx}{0.85\textwidth}{|c|cYY|YY|YY|YY|}
			\hline
			&  & \multicolumn{2}{c|}{\textbf{PCA Dip}} & \multicolumn{2}{c|}{\textbf{PCA Silv}}  & \multicolumn{2}{c|}{\textbf{SPCA Dip}} &  \multicolumn{2}{c|}{\textbf{SPCA Silv}} \\
			\cline{2-10}
			\multicolumn{1}{|c|}{\multirow{-2}{*}{\centering Data set}}  &  & \textcolor{dr}{\textbf{ODS}} & \textcolor{db}{\textbf{CRDS}} & \textcolor{dr}{\textbf{ODS}} & \textcolor{db}{\textbf{CRDS}} & \textcolor{dr}{\textbf{ODS}} & \textcolor{db}{\textbf{CRDS}}  & \textcolor{dr}{\textbf{ODS}} & \textcolor{db}{\textbf{CRDS}} \\
			\textbf{Ls} & & \cellcolor{dr}\textbf{0}       & \cellcolor{lb}\textcolor{white}{\textbf{0.09636}} & \cellcolor{dr}\textbf{0}       & 0.00478          & \cellcolor{dr}\textbf{0}    & \cellcolor{lb}\textcolor{white}{\textbf{0.0958}} & \cellcolor{dr}\textbf{0}          & 0.00478 \\
			\textbf{Lc} & & 0.78035 & 0 & 0.08012  & 0 & 0.67689 &	0 &	0.08486 &	0 \\
			\textbf{So} & & \cellcolor{lr}\textbf{0.00032} & \cellcolor{db}\textcolor{white}{\textbf{0.86359}} & \cellcolor{lr}\textbf{0.00291} & \cellcolor{db}\textcolor{white}{\textbf{0.28884}} & \cellcolor{lr}\textbf{0.00031}       & \cellcolor{db}\textcolor{white}{\textbf{0.86425}} & \cellcolor{lr}\textbf{0.00248}       & \cellcolor{db}\textcolor{white}{\textbf{0.28313}} \\
			\textbf{Zo} & & \cellcolor{dr}\textbf{0}       & \cellcolor{db}\textcolor{white}{\textbf{0.65651}} & \cellcolor{lr}\textbf{0.00001} & \cellcolor{lb}\textcolor{white}{\textbf{0.04144}} & \cellcolor{dr}\textbf{0} & \cellcolor{db}\textcolor{white}{\textbf{0.66729}} &  \cellcolor{lr}\textbf{0.00001}     & \cellcolor{lb}\textcolor{white}{\textbf{0.04497}}   \\		
			\textbf{Ps} & & 0.93718          & \cellcolor{db}\textcolor{white}{\textbf{0.75780}} & 0.06419          & \cellcolor{lb}\textcolor{white}{\textbf{0.07279}} & 0.93472 & \cellcolor{db}\textcolor{white}{\textbf{0.71148}} & 0.07014     & \cellcolor{lb}\textcolor{white}{\textbf{0.03648}} \\
			\textbf{Hr} & & \cellcolor{lr}\textbf{0.00150} & \cellcolor{lb}\textcolor{white}{\textbf{0.03202}} & \cellcolor{lr}\textbf{0.00057} & \cellcolor{lb}\textcolor{white}{\textbf{0.02910}} & \cellcolor{lr}\textbf{0.00151} & \cellcolor{lb}\textcolor{white}{\textbf{0.03207}} & \cellcolor{lr}\textbf{0.00101} & \cellcolor{lb}\textcolor{white}{\textbf{0.02324}} \\
			\textbf{Ly} & & 0.55321   & 0   & 0.59696    & 0   & 0.54266 & 0   & 0.59236   & 0   \\
			\textbf{Hd} & & 0.60077          & 0                & \cellcolor{lr}\textbf{0.00200} & 0  & 0.58349 & 0 & \cellcolor{lr}\textbf{0.00176}       & 0  \\
			\textbf{Sf} & & \cellcolor{lr}\textbf{0.00292} & 0  & 0.03277   & 0 &  \cellcolor{lr}\textbf{0.00292}  & 0 & 0.03410  & 0   \\		
			\textbf{Pt} & & 0.30036  & 0  & 0.10483   & 0   & 0.30620 & 0 & 0.15339  & 0  \\
			\textbf{De} & & \cellcolor{lr}\textbf{2E-06}   & \cellcolor{db}\textcolor{white}{\textbf{0.88778}} & \cellcolor{dr}\textbf{0}  & \cellcolor{db}\textcolor{white}{\textbf{0.24010}} & \cellcolor{dr}\textbf{7E-07} & \cellcolor{db}\textcolor{white}{\textbf{0.91679}} & \cellcolor{dr}\textbf{0}   &  \cellcolor{db}\textcolor{white}{\textbf{0.23909}}  \\
			\textbf{Hv} & & \cellcolor{dr}\textbf{0}       & \cellcolor{db}\textcolor{white}{\textbf{0.37398}} & \cellcolor{dr}\textbf{0}       & 0.00416          & \cellcolor{dr}\textbf{0} & \cellcolor{db}\textcolor{white}{\textbf{0.38693}} & \cellcolor{dr}\textbf{0}  & 0.00101 \\
			\textbf{Bs} & & \cellcolor{dr}\textbf{0}       & \cellcolor{lb}\textcolor{white}{\textbf{0.09860}} & \cellcolor{dr}\textbf{0}       & \cellcolor{lb}\textcolor{white}{\textbf{0.02363}} & \cellcolor{dr}\textbf{0}       & \cellcolor{lb}\textcolor{white}{\textbf{0.10093}} & \cellcolor{dr}\textbf{0}       & \cellcolor{lb}\textcolor{white}{\textbf{0.02287}} \\
			\textbf{Ca} & & \cellcolor{dr}\textbf{0}       & 0.00001          & \cellcolor{dr}\textbf{0}       & \cellcolor{lb}\textcolor{white}{\textbf{0.06579}} & \cellcolor{dr}\textbf{0} & 0.00001 & \cellcolor{dr}\textbf{0} & \cellcolor{lb}\textcolor{white}{\textbf{0.06274}}  \\
			\textbf{Bc} & & \cellcolor{lr}\textbf{0.00001} & 0  & \cellcolor{dr}\textbf{0}  & 0  & \cellcolor{lr}\textbf{0.00001} & 0 & \cellcolor{dr}\textbf{0}  & 0  \\
			\textbf{Mm} & & \cellcolor{dr}\textbf{0} & 0    & \cellcolor{dr}\textbf{0}  & 0   & \cellcolor{dr}\textbf{0} & 0 & \cellcolor{dr}\textbf{0} & 0  \\
			\textbf{Tt} & & \cellcolor{lr}\textbf{0.00334} & 0  & \cellcolor{dr}\textbf{0} & 0 & \cellcolor{lr}\textbf{0.00312} & 0 & \cellcolor{lr}\textbf{0.00001}  & 0   \\
			\textbf{Ce} & & \cellcolor{dr}\textbf{0}  & 0   & \cellcolor{dr}\textbf{0}  & 0  & \cellcolor{dr}\textbf{0}  & 0 & \cellcolor{dr}\textbf{0}  & 0   \\
			\hline
		\end{tabularx}
\end{table}

\begin{table}[h]
	\caption*{\textbf{Supplementary Table 3.2}: The results of clusterability evaluation methods utilizing \textbf{CDE} on 18 UCI data sets.} 
	\centering
	\begin{tabularx}{0.85\textwidth}{|c|cYY|YY|YY|YY|}
		\hline
		&  & \multicolumn{2}{c|}{\textbf{PCA Dip}} & \multicolumn{2}{c|}{\textbf{PCA Silv}}  & \multicolumn{2}{c|}{\textbf{SPCA Dip}} &  \multicolumn{2}{c|}{\textbf{SPCA Silv}} \\
		\cline{2-10}
		\multicolumn{1}{|c|}{\multirow{-2}{*}{\centering Data set}}  &  & \textcolor{dr}{\textbf{ODS}} & \textcolor{db}{\textbf{CRDS}} & \textcolor{dr}{\textbf{ODS}} & \textcolor{db}{\textbf{CRDS}} & \textcolor{dr}{\textbf{ODS}} & \textcolor{db}{\textbf{CRDS}}  & \textcolor{dr}{\textbf{ODS}} & \textcolor{db}{\textbf{CRDS}} \\
		\textbf{Ls}  & & \cellcolor{lr}\textbf{0.00170}  & \cellcolor{db}\textcolor{white}{\textbf{0.55578}} & \cellcolor{dr}\textbf{0}        & \cellcolor{db}\textcolor{white}{\textbf{0.52003}} & \cellcolor{lr}\textbf{0.00170}  & \cellcolor{db}\textcolor{white}{\textbf{0.55368}} & 0.09006   & \cellcolor{db}\textcolor{white}{\textbf{0.47639}} \\	 		
		\textbf{Lc}  & & 0.92497  & \cellcolor{db}\textcolor{white}{\textbf{0.33582}} & 0.35462  & 0.00078  & 0.92700  & \cellcolor{db}\textcolor{white}{\textbf{0.34050}}  & 0.35577 & 0.00038 \\
		\textbf{So}  & & \cellcolor{lr}\textbf{0.00401}  & \cellcolor{db}\textcolor{white}{\textbf{0.77441}} & 0.15443  & \cellcolor{db}\textcolor{white}{\textbf{0.36268}} & \cellcolor{lr}\textbf{0.00399} & \cellcolor{db}\textcolor{white}{\textbf{0.76650}} & 0.10749  & \cellcolor{db}\textcolor{white}{\textbf{0.39486}} \\
		\textbf{Zo}  & & \cellcolor{dr}\textbf{0}        & \cellcolor{lb}\textcolor{white}{\textbf{0.08946}} & \cellcolor{dr}\textbf{0}  & \cellcolor{lb}\textcolor{white}{\textbf{0.02910}} & \cellcolor{dr}\textbf{0}  & \cellcolor{lb}\textcolor{white}{\textbf{0.09142}} & \cellcolor{dr}\textbf{0}  & \cellcolor{lb}\textcolor{white}{\textbf{0.02143 }}   \\
		\textbf{Ps}  & & 0.99092           & \cellcolor{db}\textcolor{white}{\textbf{0.71008}} & 0.77838           & \cellcolor{db}\textcolor{white}{\textbf{0.33734}} & 0.99022  & \cellcolor{db}\textcolor{white}{\textbf{0.73436}} & 0.76116   & \cellcolor{db}\textcolor{white}{\textbf{0.31540}}  \\
		\textbf{Hr}  & & \cellcolor{lr}\textbf{0.00005} & 0.00003   & \cellcolor{dr}\textbf{0}  & 0  & \cellcolor{lr}\textbf{0.00005}  & 0.00004 & \cellcolor{lr}\textbf{0.00002} & 0 \\
		\textbf{Ly}  & & 0.08645           & \cellcolor{db}\textcolor{white}{\textbf{0.94672}} & \cellcolor{lr}\textbf{0.00002} & \cellcolor{db}\textcolor{white}{\textbf{0.80823}} & 0.08558 & \cellcolor{db}\textcolor{white}{\textbf{0.94339}} & \cellcolor{lr}\textbf{0.00001}  &  \cellcolor{db}\textcolor{white}{\textbf{0.82890}}   \\
		\textbf{Hd}  & & 0.32479           & \cellcolor{db}\textcolor{white}{\textbf{0.76356}} & 0.02453           & \cellcolor{db}\textcolor{white}{\textbf{0.85990}} & 0.32455 & \cellcolor{db}\textcolor{white}{\textbf{0.75964}}  & 0.02505  & \cellcolor{db}\textcolor{white}{\textbf{0.86335}}  \\
		\textbf{Sf}  & & \cellcolor{dr}\textbf{0} & 0.00016  & \cellcolor{dr}\textbf{0}  & 0  & \cellcolor{dr}\textbf{0} & 0.00015 & \cellcolor{dr}\textbf{0} & 0.00057 \\
		\textbf{Pt}  & & 0.20672           & \cellcolor{db}\textcolor{white}{\textbf{0.61520}} & \cellcolor{lr}\textbf{0.00718}  & \cellcolor{db}\textcolor{white}{\textbf{0.13530}} & 0.20652 & \cellcolor{db}\textcolor{white}{\textbf{0.59953}} & \cellcolor{lr}\textbf{0.00542}  & \cellcolor{db}\textcolor{white}{\textbf{0.15548}} \\
		\textbf{De}  & & \cellcolor{dr}\textbf{0}        & \cellcolor{db}\textcolor{white}{\textbf{0.90482}} & \cellcolor{dr}\textbf{0} & \cellcolor{db}\textcolor{white}{\textbf{0.20639}} & \cellcolor{dr}\textbf{0} & \cellcolor{db}\textcolor{white}{\textbf{0.89693}} & \cellcolor{dr}\textbf{0} & \cellcolor{db}\textcolor{white}{\textbf{0.20333}}   \\
		\textbf{Hv}  & & \cellcolor{dr}\textbf{0} & 0  & \cellcolor{dr}\textbf{0}  & 0  & \cellcolor{dr}\textbf{0} & 0 & \cellcolor{dr}\textbf{0}  & 0 \\
		\textbf{Bs}  & & \cellcolor{dr}\textbf{0}        & 0.00643          & \cellcolor{dr}\textbf{0}        & 0  & \cellcolor{dr}\textbf{0} & 0.00614 & \cellcolor{dr}\textbf{0}  & 0 \\
		\textbf{Ca}  & & \cellcolor{dr}\textbf{0} & 0  & \cellcolor{dr}\textbf{0} & 0 &  \cellcolor{dr}\textbf{0} & 0 & \cellcolor{dr}\textbf{0} & 0 \\
		\textbf{Bc}  & & \cellcolor{dr}\textbf{0}  & 0   & \cellcolor{dr}\textbf{0}  & 0  & \cellcolor{dr}\textbf{0} & 0 & \cellcolor{dr}\textbf{0} & 0\\
		\textbf{Mm}  & & \cellcolor{dr}\textbf{0} & 0  & \cellcolor{dr}\textbf{0}  & 0  & \cellcolor{dr}\textbf{0} & 0 & \cellcolor{dr}\textbf{0} & 0 \\
		\textbf{Tt}  & & \cellcolor{dr}\textbf{0}   & 0  & \cellcolor{dr}\textbf{0} & 0  & \cellcolor{dr}\textbf{0} & 0 & \cellcolor{dr}\textbf{0} & 0 \\
		\textbf{Ce}  & & \cellcolor{dr}\textbf{0}  & 0 & \cellcolor{dr}\textbf{0}  & 0 & \cellcolor{dr}\textbf{0} & 0 & \cellcolor{dr}\textbf{0} & 0 \\
		\hline
	\end{tabularx}
\end{table}

\begin{table}[h]
	\caption*{\textbf{Supplementary Table 4}: The adjusted $p$-values via the Monte Carlo method for each data set, using the same ODS and representative CRDS as in Table 2.}
	\centering
	\begin{tabularx}{0.5\textwidth}{|Y|Y|Y|}
		\hline
		Data set & \textbf{ODS} & \textbf{CRDS} \\
		\hline
		Ls   & 1               & 0.83   \\
		Lc   & 1.1E-36         & 1      \\
		So   & 7.6E-149        & 0.98   \\
		Zo   & 4.7E-122        & 0.89   \\
		Ps   & 1.5E-24         & 0.56   \\
		Hr   & 3.9E-05         & 0.47   \\
		Ly   & 2.2E-106        & 0.74   \\
		Hd   & 7.1E-40         & 0.35   \\
		Sf   & 3.2E-55         & 0.61   \\
		Pt   & 1.45E-57        & 0.96   \\
		De   & 0               & 0.73   \\
		Hv   & 1.41E-310       & 0.53   \\
		Bs   & 1               & 0.45   \\
		Ca   & 4.7E-63         & 0.94   \\
		Bc   & 8.84E-106       & 0.63   \\
		Mm   & 1.00E-11        & 0.63   \\
		Tt   & 4.45E-39        & 0.40   \\
		Ce   & 1               & 0.45   \\
		\hline
	\end{tabularx}
\end{table}

\FloatBarrier
\newpage
\section*{Supplementary Figures}

\begin{figure*}[h]
	\includegraphics[width=0.65\linewidth]{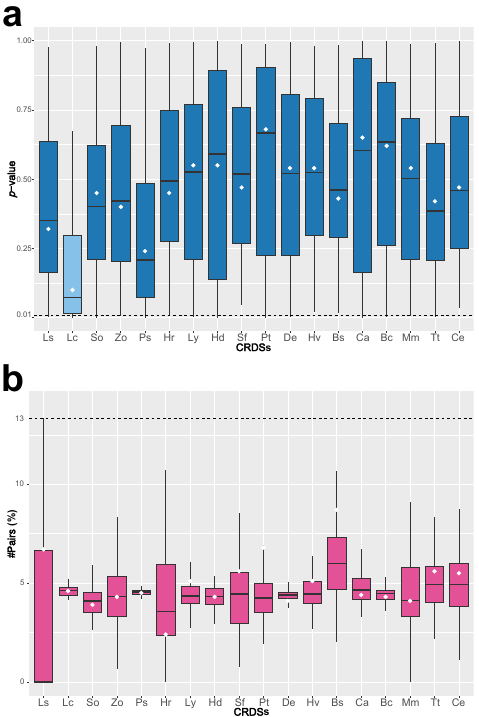} 
	\caption*{\textbf{Supplementary Figure 1}: The boxplots illustrate the distribution of \textbf{(a)} $p$-values and \textbf{(b)} proportion of strongly associated attribute pairs (calculated by TestCat) of CRDSs for each data set from \textbf{the 101 runs}. The white diamond in each box represents the $p$-value or proportion of associated attribute pairs of chosen CRDS  used across all comparison methods.} 
\end{figure*}

\begin{figure*}[t]
	\includegraphics[width=\linewidth]{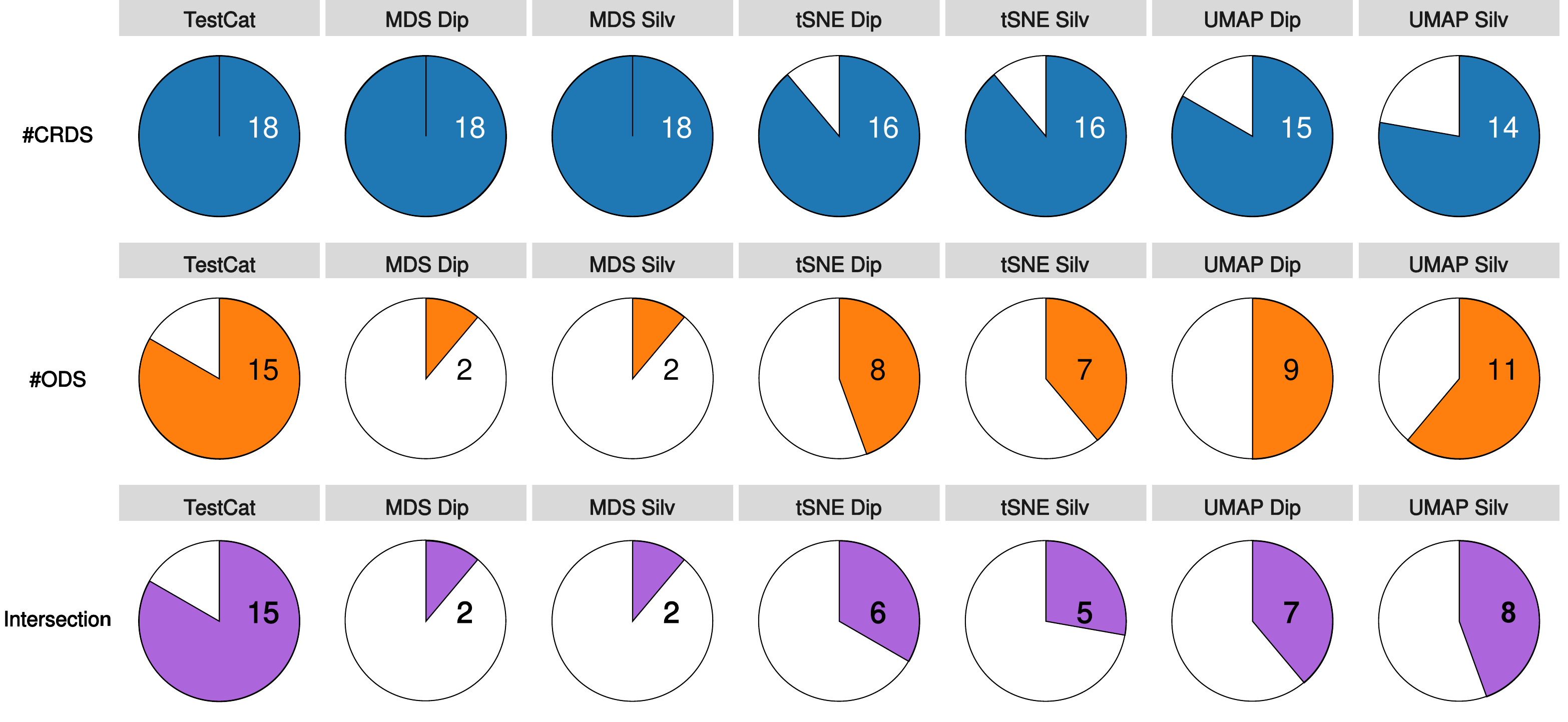} 
	\caption*{\textbf{Supplementary Figure 2}: Count of correctly identified data sets by using TestCat and compared methods via dimensionality reduction (running 101 times for each data set) based on \textbf{Lin1} distance. The resulting median $p$-value from 101 runs is used for quantifying the clusterability of each target data set.} 
\end{figure*}

\begin{figure*}[t]
	\includegraphics[width=0.755\linewidth]{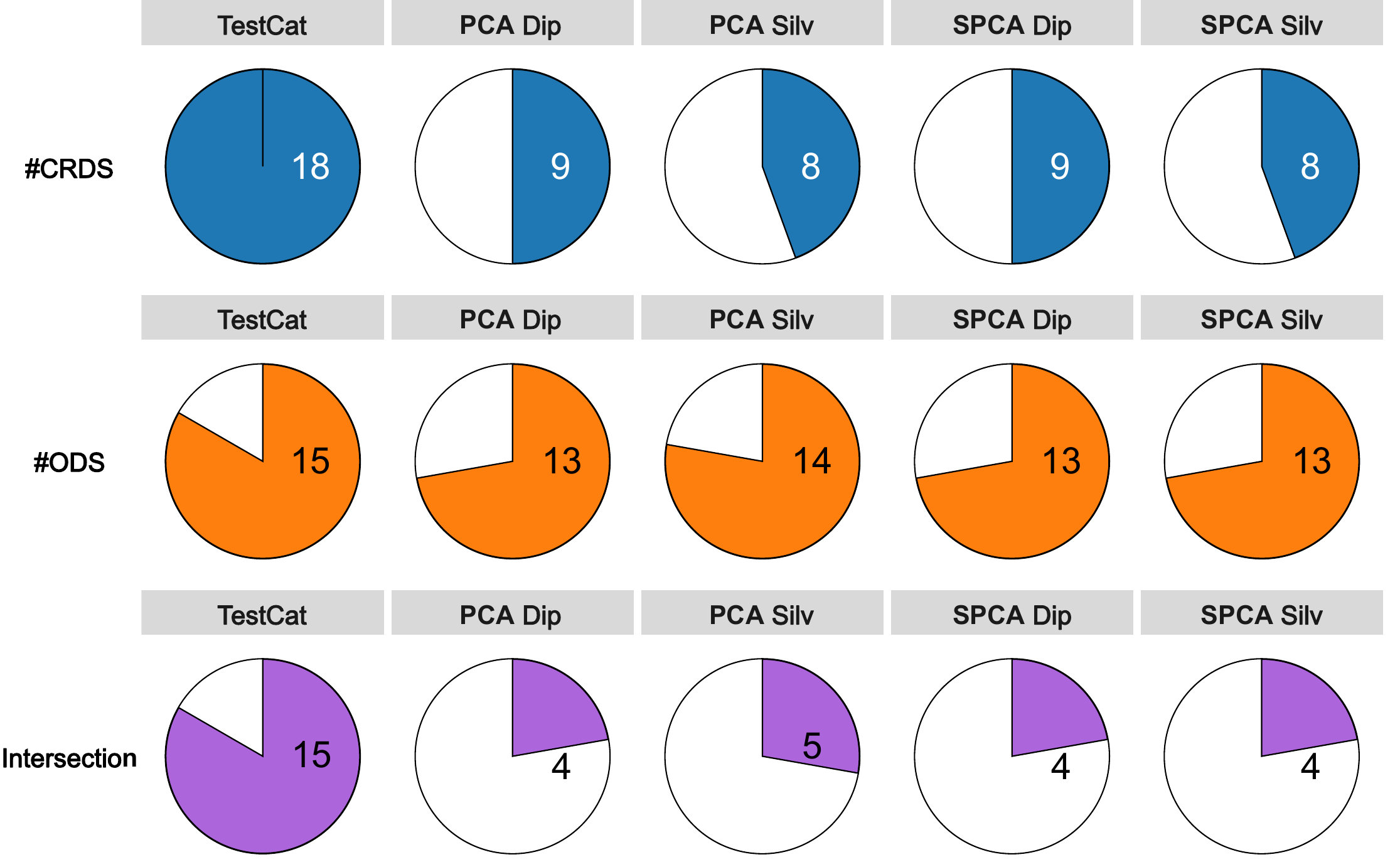} 
	\caption*{\textbf{Supplementary Figure 3}: Count of correctly identified data sets by using TestCat and Categorical-to-Numerical methods via CDE embedding. We implemented the CDE embedding on each categorical data set to generate its numerical representation. Following this transformation, we utilized PCA or SPCA to further condense this numerical data.} 
\end{figure*}

\begin{figure}[t]
	\includegraphics[width=1\linewidth]{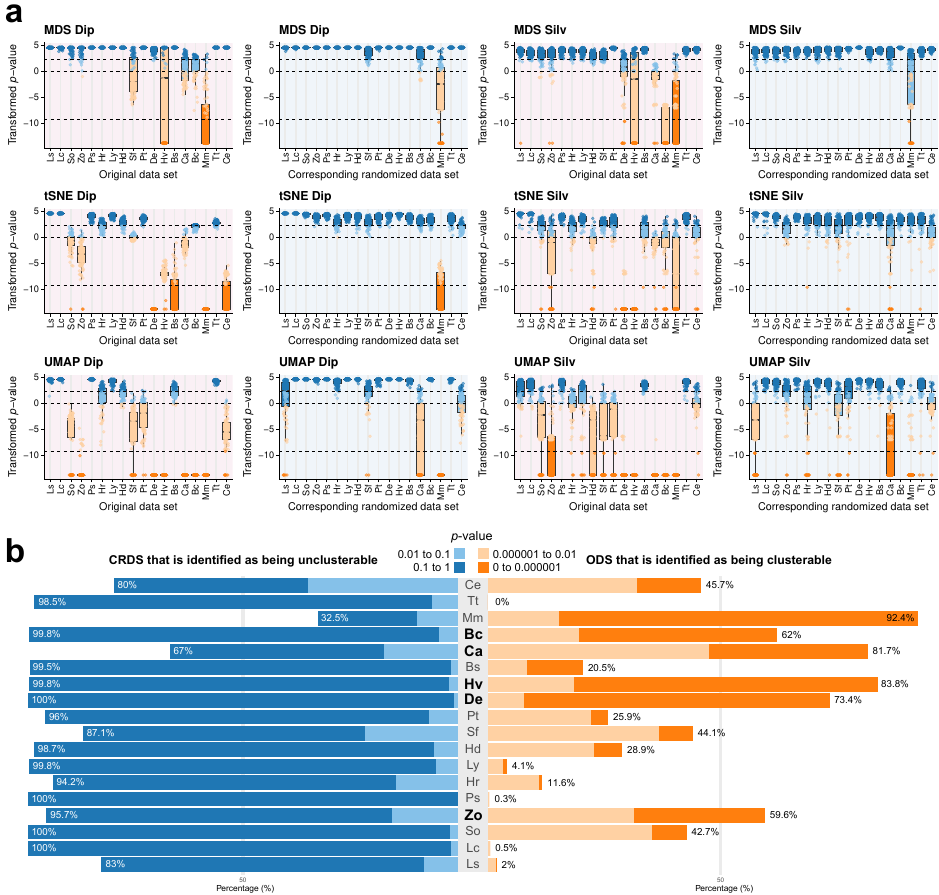} 
	\caption*{\textbf{Supplementary Figure 4.1}: Detailed identification results of Dip and Silverman via dimensionality reduction (running MDS/ tSNE/ UMAP 101 times for each data set) based on \textbf{Hamming} distance.  \textbf{(a)} The boxplots of $p$-values on 18 UCI data sets (including original data sets and corresponding randomized data sets) are produced by using the listed methods. To better visualize smaller $p$-values that approach or equal 0, and to distinguish the significance level of 0.01 through y-axis ($y=0$), we use transformed $p$-values, defined by the formula: $y=\log (p\text{-value}/ 0.01 + 0.000001)-\log (0.000001)$. Boxes and scatter points, representing overall experimental results and individual outcomes respectively, are colored based on their $p$-value category as indicated in the legend. The color of each box is determined by the median number. \textbf{(b)} The stacked barplots provide a summary of the overall performance from 101*6 $p$-values by using 6 methods (MDS Dip, MDS Silv, tSNE Dip, tSNE Silv, UMAP Dip, UMAP Silv), separately for original data sets and corresponding randomized data sets. The two groups of stacked barplots count the ratio of reported unclusterable CRDS and clusterable ODS. Each data set where these methods yield a ratio over $50\%$ with respect to both  CRDS and ODS is highlighted in bold. The experimental results based on Lin1 are displayed in Supplementary Figure 4.2.} 
\end{figure}

\begin{figure*}[t]
	\includegraphics[width=1\linewidth]{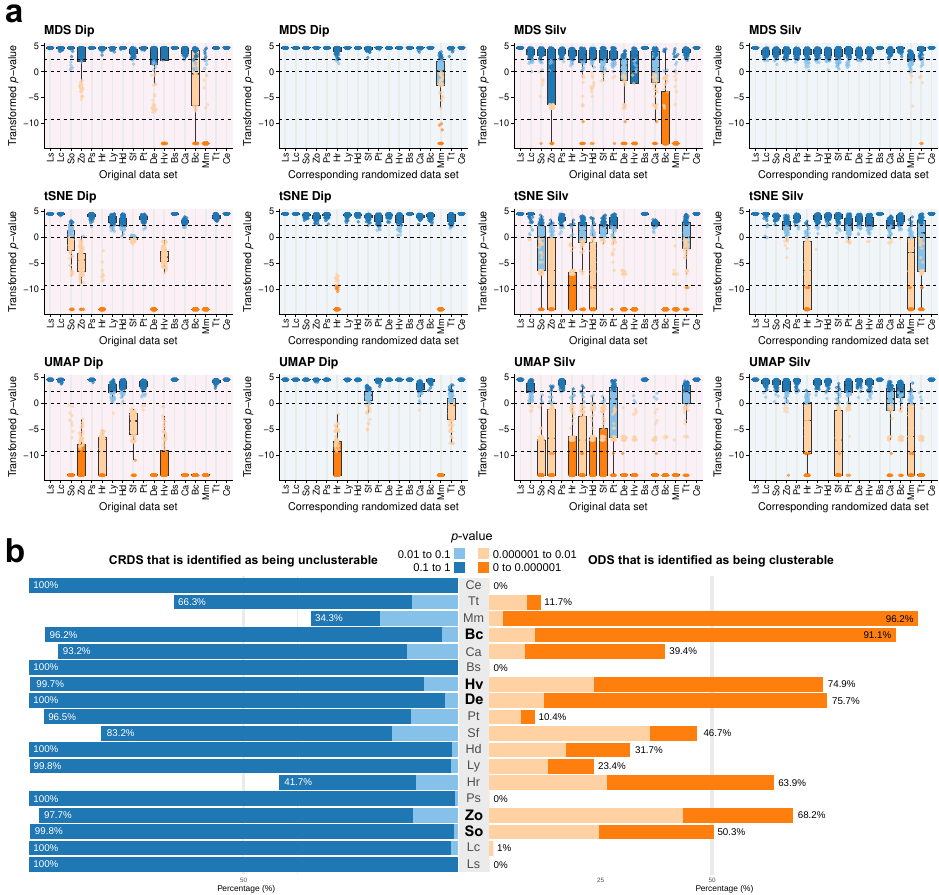} 
	\caption*{\textbf{Supplementary Figure 4.2}: Detailed experimental results of Dip and Silverman via dimensionality reduction (running 101 times for each data set) based on \textbf{Lin1} distance. The meaning of \textbf{(a)} and \textbf{(b)} are the same as those in Supplementary Figure 4.1.} 
\end{figure*}

\begin{figure*}[h]
	\includegraphics[width=0.95\linewidth]{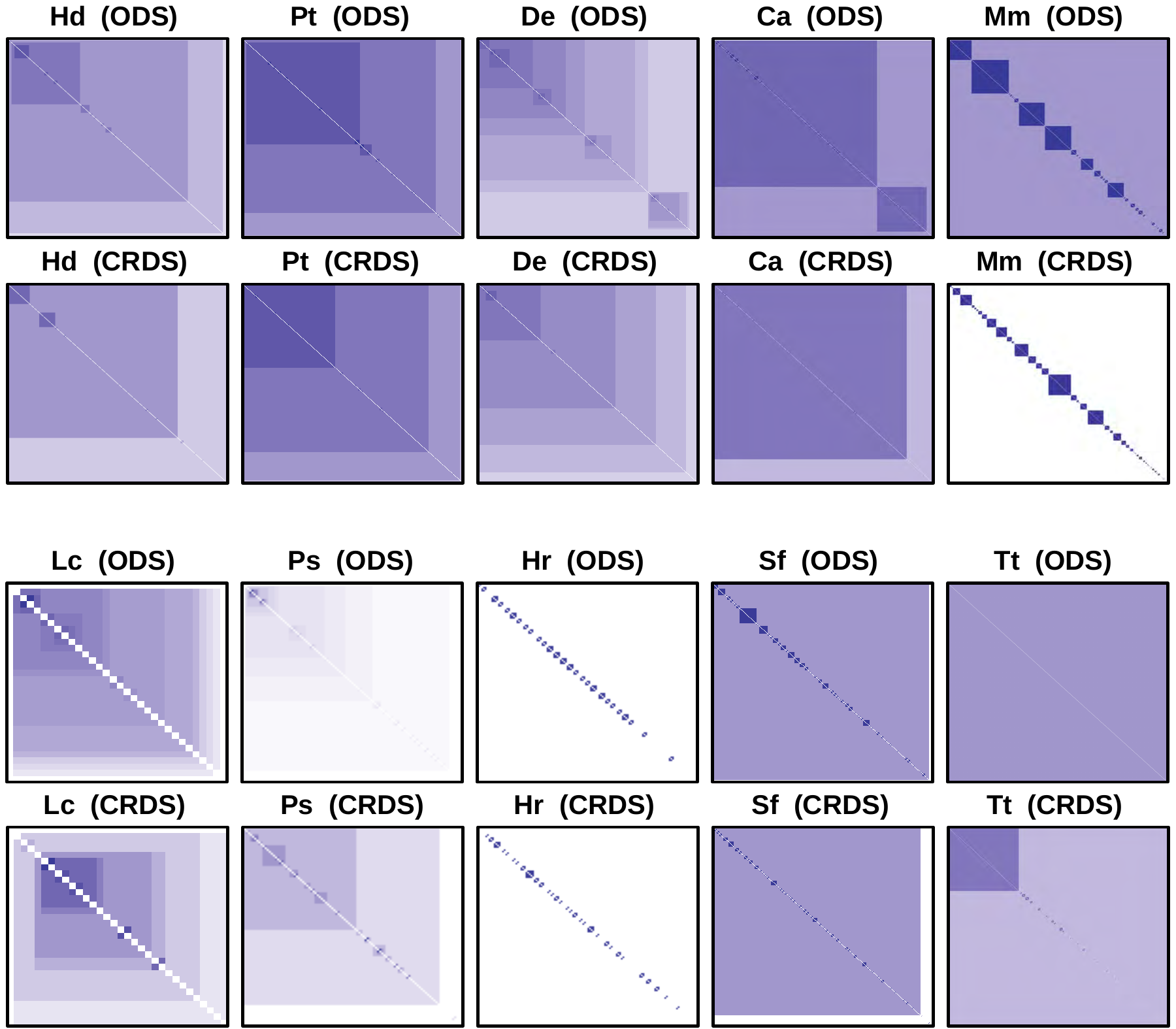} 
	\caption*{\textbf{Supplementary Figure 5}: iVAT plots of 10 UCI categorical data sets. All plots are derived from the Hamming distances between objects in the data sets.} 
\end{figure*}

\begin{figure*}[t]
	\includegraphics[width=1\linewidth]{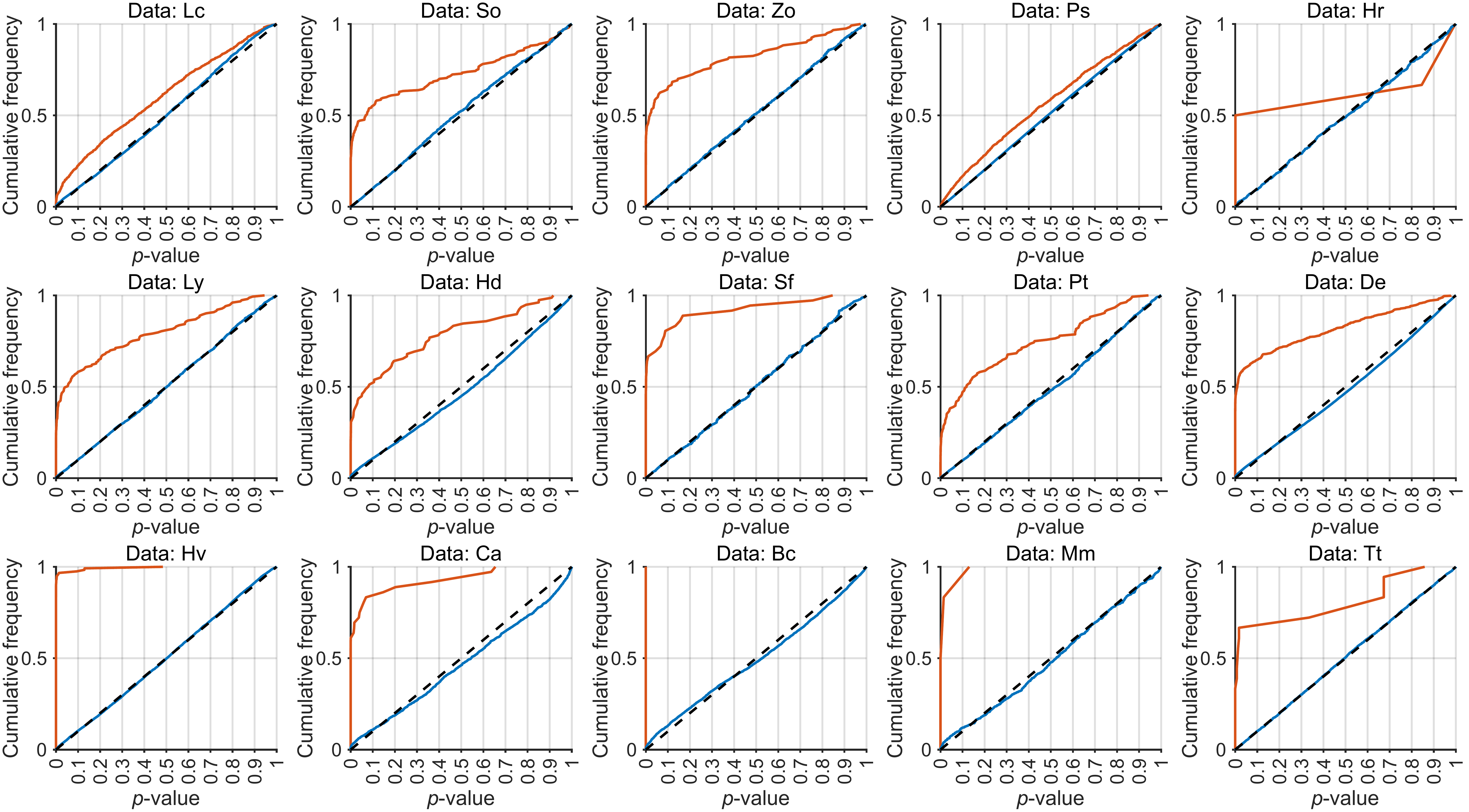} 
	\caption*{\textbf{Supplementary Figure 6}: Comparison of empirical CDFs for $p$-values of attribute pairs across 15 data sets with a Uniform CDF. The orange lines represent the same meaning as in Figure 6, while the blue line for each plot is derived from 101 independently generated CRDSs based on ODS, providing sufficient samples for clearer observation.}
\end{figure*}

\begin{figure*}[t]
	\includegraphics[width=0.85\linewidth]{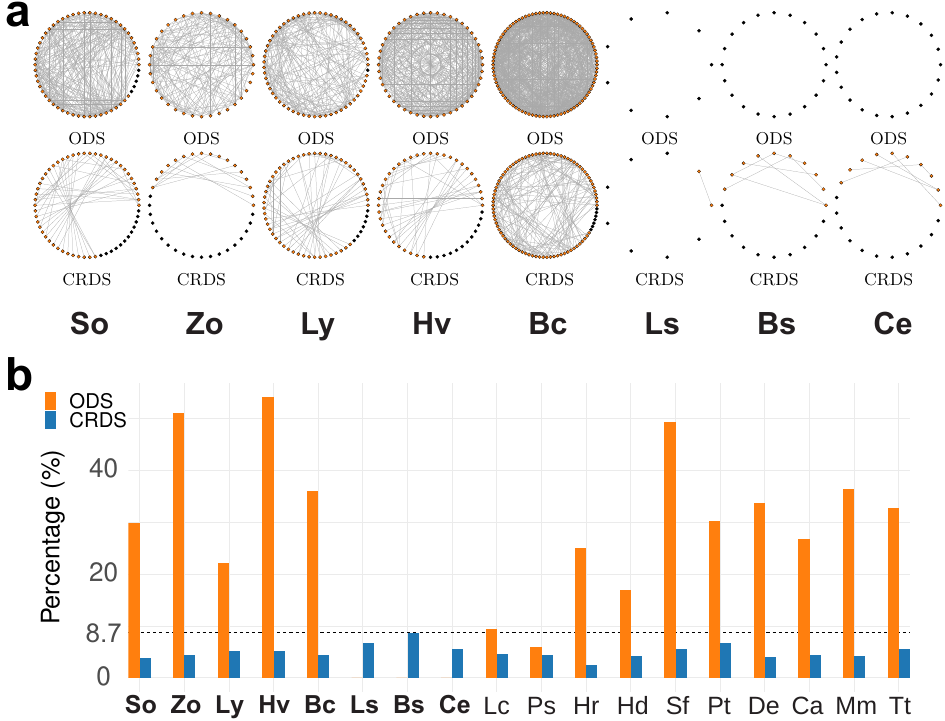} 
	\caption*{\textbf{Supplementary Figure 7}: The number of positively associated attribute value pairs and the proportion of strongly associated attribute pairs in ODS and  CRDS. \textbf{(a)} Two attribute values are regarded to be positively associated and connected with a gray line when their standardized residuals exceed 2. The black dots indicate that the attribute values have no strong positive association with others. The plots on other ten data sets are displayed in Supplementary Figure 7. \textbf{(b)} Comparison of the proportion of strongly associated attribute pairs between ODS and the representative CRDS for 18 data sets.} 
	\label{fig:nets}
\end{figure*}

\begin{figure*}[h]
	\includegraphics[width=1\linewidth]{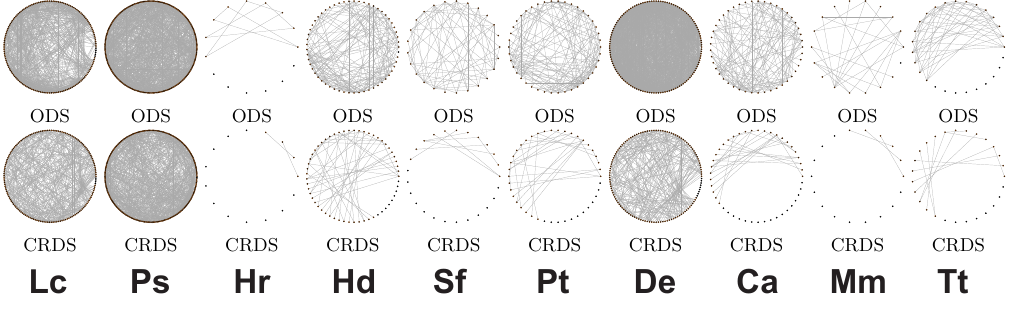} 
	\caption*{\textbf{Supplementary Figure 8}: The number of positively associated attribute value pairs for the ODS and CRDS of ten data sets. The meaning of the plots are the same as those in Supplementary Figure 6a.} 
\end{figure*}